\documentclass[10pt,twocolumn,letterpaper]{article}

\usepackage[pagenumbers]{cvpr} 

\usepackage[utf8]{inputenc} 
\usepackage[T1]{fontenc}    
\usepackage{url}            
\usepackage{booktabs}       
\usepackage{amsfonts}       
\usepackage{nicefrac}       
\usepackage{microtype}      
\usepackage{xcolor}         
\usepackage{subcaption}
\usepackage{graphicx}
\usepackage{booktabs} 
\usepackage{caption} 
\usepackage{multirow} 
\usepackage{enumitem}
\usepackage{tcolorbox} 
\usepackage{xcolor}    
\usepackage{listings}

\usepackage[accsupp]{axessibility} 
\definecolor{lightyellow}{rgb}{1, 0.95, 0.85}
\definecolor{graphicbackground}{rgb}{0.9765,0.9451,0.9059}
\definecolor{codebackground}{rgb}{0.8314,0.949,0.9882}

\definecolor{cvprblue}{rgb}{0.21,0.49,0.74}
\usepackage[pagebackref,breaklinks,colorlinks,allcolors=cvprblue]{hyperref}

\def\paperID{15372} 
\def\confName{CVPR}
\def\confYear{2025}

\title{Are Images Indistinguishable to Humans Also Indistinguishable to Classifiers?}
\author{Zebin You$^{1,2}$\thanks{Work done during an internship at Baidu VIS.}, Xinyu Zhang$^{3}$, Hanzhong Guo$^{1,2}$, Jingdong Wang$^{4}$, Chongxuan Li$^{1,2}$\thanks{Correspondence to Chongxuan Li.} \\
  $^1$ Gaoling School of Artificial Intelligence, Renmin University of China, Beijing, China \\
  $^2$ Beijing Key Laboratory of Big Data Management and Analysis Methods, Beijing, China \\
  $^3$ The University of Adelaide
  $^4$ Baidu VIS \\
  \texttt{\{zebin, guohanzhong, chongxuanli\}@ruc.edu.cn;} \\
  \texttt{xyzhang0717@gmail.com;} \texttt{wangjingdong@baidu.com}
}

\begin{document}
\maketitle
\begin{abstract}
The ultimate goal of generative models is to perfectly capture the data distribution. For image generation, common metrics of visual quality (e.g., FID) and the perceived truthfulness of generated images seem to suggest that we are nearing this goal. However, through distribution classification tasks, we reveal that, from the perspective of neural network-based classifiers, even advanced diffusion models are still far from this goal. Specifically, classifiers are able to consistently and effortlessly distinguish real images from generated ones across various settings. Moreover, we uncover an intriguing discrepancy: classifiers can easily differentiate between diffusion models with comparable performance (e.g., U-ViT-H vs. DiT-XL), but struggle to distinguish between models within the same family but of different scales (e.g., EDM2-XS vs. EDM2-XXL). Our methodology carries several important implications. First, it naturally serves as a diagnostic tool for diffusion models by analyzing specific features of generated data. Second, it sheds light on the model autophagy disorder and offers insights into the use of generated data: augmenting real data with generated data is more effective than replacing it. Third, classifier guidance can significantly enhance the realism of generated images.
\end{abstract}    
\section{Introduction}
\label{sec:intro}
\begin{figure}[t]
    \centering
    \includegraphics[width=\linewidth]{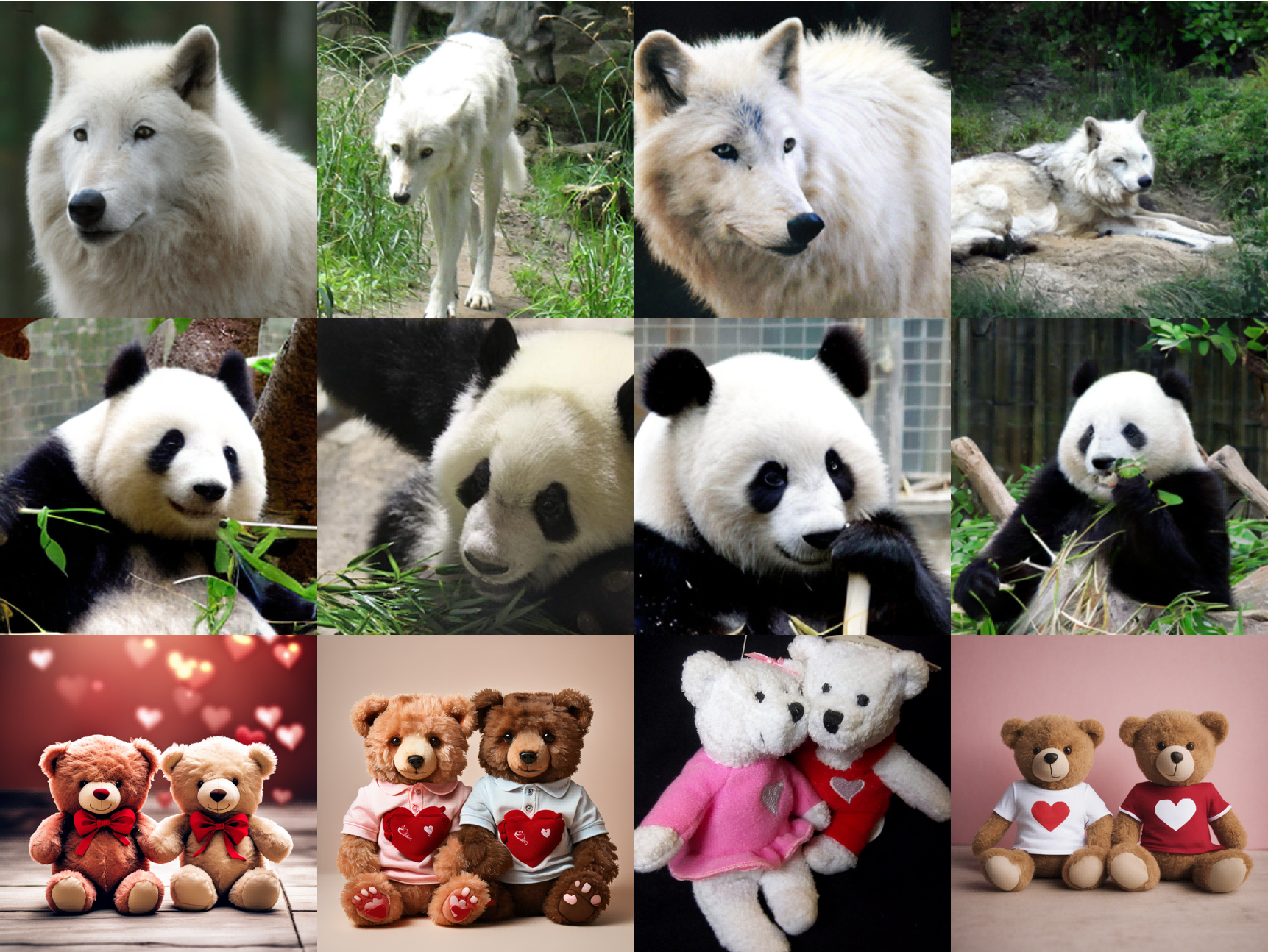}
    \caption{\textbf{Four-way distribution classification tasks:} \emph{Which one is real in each row?}\protect\footnotemark The samples are from real images or generated from state-of-the-art diffusion models. Notably, classifiers consistently and effortlessly distinguish between real and generated images in all settings.}
    \label{fig:name-that-true}
\end{figure}

Diffusion probabilistic models~\citep{sohl2015deep, song2020score, ho2020denoising} have emerged as leading generative models for image~\citep{ramesh2021zero, esser2024scaling}, video~\citep{videoworldsimulators2024, bao2024vidu, zhao2024identifying} and 3D content generation~\citep{poole2022dreamfusion}, surpassing previous generative models such as generative adversarial networks~\citep{goodfellow2014generative}, variational autoencoders~\citep{kingma2013auto}, and normalizing flows~\citep{rezende2015variational}. In particular, focusing on image generation, diffusion models are capable of producing realistic images that are often indistinguishable from real ones to the human eye (see Fig.~\ref{fig:name-that-true}). Furthermore, common metrics such as FID~\citep{heusel2017gans} suggest that the distribution generated by state-of-the-art diffusion models closely approximates the ImageNet validation set~\citep{karras2023analyzing, karras2022elucidating, tian2024visual}. Therefore some works~\cite{wang2024generated, you2024diffusion}, have utilized generated data to augment training datasets. However, to the best of our knowledge, few studies have been conducted to analyze the differences between the generated and real distributions.

\footnotetext{From left to right, the first two rows show images from U-ViT, ImageNet~(real), DiT, and EDM2-XXL. The bottom row shows images from Pixart-$\alpha$, Playground-v2.5, COCO~(real), and SDXL.}

This raises a fundamental question: \textit{How far is the generated distribution from the real distribution?} Drawing inspiration from methodologies in dataset bias~\citep{liu2024decade, tommasi2017deeper, torralba2011unbiased}, we propose distribution classification tasks (see Sec.~\ref{sec:related_work} for details) that predict whether an image belongs to the real or generated distribution. These tasks serve as an effective probe to quantify and analyze this distributional gap. We use the term ``distribution classification'' because generative models aim to capture an entire distribution rather than a finite dataset, allowing infinite sampling. 

Notably, binary classification is a natural approach, as classifiers parameterized by neural networks can serve as effective indicators of the differences between the real and generated distributions through classification accuracy (see more analysis in Appendix ~\ref{app:binary_classification_distance}). Compared to the widely used FID metric, our classifier-based approach directly measures distribution differences without relying on Gaussian assumptions (see further discussion in Appendix~\ref{app:relationship_fid}). 

We first explore \textit{are images indistinguishable to humans also indistinguishable to classifiers?} To this end, we train several commonly used classifiers to distinguish between the real and generated distribution across various settings. Quantitatively, most classifiers can achieve over 98\% accuracy. Meanwhile, all classifiers reach extremely high accuracy with significantly fewer training samples than the requirement of training diffusion models. Interestingly, even self-supervised classifiers exhibit this capacity to identify the generated distribution. These results demonstrate that from the perspective of neural network-based classifiers, the generated distribution remains markedly different from the real one, indicating that diffusion models still have significant room for improvement.

Furthermore, we observe intriguing contradictions between classifiers' prediction and both FID and human judgments through two groups of experiments. The first group involves classifying the smallest and largest models within the same diffusion family, which share inductive biases but differ in visual quality. The second group examines diffusion models with similar FID and visual quality but different inductive biases. Surprisingly, classifiers perform well on the second task but struggle with the first, revealing contradictions between classifier performance and FID. Meanwhile, human judgments align with classifiers in the first task but degrade to random guessing in the second, indicating inconsistency between classifier and human judgments.

Our approach displays several important implications. First, it naturally serves as a diagnostic tool for diffusion models by extending the analysis to specific features of generated data. For example, in evaluating the advanced diffusion transformer U-ViT-H/2~\citep{bao2023all}, we focus on two key aspects: spatial features and frequency ranges. Spatially, classifiers can easily distinguish whether an image is real or generated using any part of the image. In terms of frequency, classifiers struggle with minimal low-frequency information but perform well across all other frequency intervals. This suggests that U-ViT-H/2 has effectively learned low-frequency features, making generated images harder to distinguish in these regions, 
but remains less proficient in higher-frequency regions.
Second, our approach enables the evaluation of using generated data for training models. It provides a reasonable explanation for the model autophagy disorder (MAD) phenomenon observed in previous work~\cite{shumailov2024ai, alemohammad2024self}. This collapse is surprising because, despite the realistic appearance of the generated data, using it in a continuous loop to train subsequent generations of models leads to failure. We explain this by showing that there exists a distribution mismatch between the training and generated distributions in each generation. As this mismatch accumulates, the generated data distribution drifts further from the original. In downstream tasks, such as supervised and semi-supervised classification, this mismatch leads to performance degradation when real data is simply replaced by generated data. 
However, this distribution drift can be mitigated by \emph{augmenting} real data with generated data, which introduces novel features while maintaining alignment with the original data distribution, leading to improved performance.

Briefly, our key contributions are as follows:
\begin{itemize}[leftmargin=*]
\setlength\itemindent{.5em}
\item We show that classifiers easily distinguish between diffusion-generated and real distributions, despite diffusion models achieving low FID scores and generating lifelike images (see Fig.~\ref{fig:name-that-true}).
\item We reveal the intriguing contradictions between classifier performance and widely used evaluation metrics, such as FID and human judgments.
\item We demonstrate that our approach complements traditional metrics like FID and can serve as a diagnostic tool to provide deeper insights into diffusion models.
\item We provide a reasonable explanation for the model autophagy disorder phenomenon and show that augmenting real data with generated data is more effective than replacing it in supervised and semi-supervised learning.
\item We demonstrate that classifier guidance can be used to enhance the realism of generated images.
\end{itemize}

\section{Related work}
\label{sec:related_work}
\textbf{Dataset bias.} \citet{torralba2011unbiased} introduced the "Name That Dataset" game to discover dataset bias. \citet{tommasi2017deeper} and \citet{liu2024decade} expanded on this by applying convolutional neural networks and large-scale datasets, respectively. Existing work on dataset bias focuses on classification to highlight biases in data collection for building unbiased datasets, our work leverages distribution classification to explore a fundamental question: \emph{How far is the generated distribution from the real distribution?} Besides, we focus on the discrepancy between the generated distribution and the real distribution rather than between specific datasets. Furthermore, the conclusion drawn by related work highlights the ability of neural networks to detect biases within increasingly general datasets, our conclusion underscores the significant gap that still exists between current diffusion models and the real data distribution. We also highlight the potential of distribution classification methods for gaining deeper insights into generative models and more effectively leveraging generated data.

\textbf{Generated image detection.} Several empirical works on generated image detection span three key areas: universal detection~\citep{park2024community, liu2024mixture, ojha2023towards, wang2020cnn, girish2021towards}, reconstruction-based methods~\citep{wang2023dire, luo2024lare}, and frequency analysis~\citep{wang2020high, zhang2019detecting, durall2020watch, chandrasegaran2021closer, frank2020leveraging, dzanic2020fourier, yang2023fingerprints, corvi2023detection, ricker2022towards, corvi2023intriguing}. Prior research typically focuses on detecting images from various generative models trained on diverse datasets, often without complete access to the training data. Consequently, simple classification methods prove ineffective, necessitating more complex detection techniques. In contrast, our work utilizes a distinct setup where the classifier is trained with both generated and original training data, enabling effective discrimination through straightforward classification. Furthermore, our primary objective is not universal generated image detection, but rather quantifying the distributional discrepancy between real and generated data. Besides, instead of prioritizing classifier performance, we focus on the insights provided into generative model behavior and limitations.

\textbf{Training with generated images.} Several studies have explored the use of generated data to enhance classification performance~\citep{zhou2023training, azizi2023synthetic, he2022synthetic}. These works typically leverage pre-trained text-to-image diffusion models, such as Stable Diffusion~\citep{rombach2022high}. In contrast, our experiments exclusively utilize generative models trained on the original training data. Other research has investigated learning visual representations solely from synthetic images and captions~\citep{tian2024learning, fan2024scaling}, demonstrating the viability of using synthetic data and achieving strong performance even with purely generated data. Our findings indicate that augmenting real data with generated data yields better results than complete replacement.

\section{Experimental settings}
\label{sec:exp_settings}
We present the main experimental settings as follows. For more details, please see Appendix~\ref{app:exp_settings}.

\textbf{Dataset.} For label-to-image tasks, we use the CIFAR-10~\citep{cifar} and ImageNet~\citep{imagenet} datasets, and for text-to-image tasks, we utilize the COCO2014 dataset~\citep{coco}. For CIFAR-10, the real distribution is constructed from its training and testing sets, comprising 50k training images and 10k testing images. The generated distribution is created using diffusion models to produce an equivalent number of images. For ImageNet, we consider resolutions of 256 and 512 and follow ADM's~\citep{dhariwal2021diffusion} method to preprocess ImageNet into the desired resolutions. We randomly sample 100k training images and 50k validation images from the corresponding ImageNet sets to represent the real distribution, and generate an equivalent number of images using diffusion models for the generated distribution. For COCO2014, we randomly sample 10k training images and 1k validation images from the respective sets to form the real distribution, and use the corresponding prompts to generate an equivalent number of data for the generated distribution.

\textbf{Classifier.} By default, we use ResNet-50~\citep{he2016deep} as the classifier architecture, and also consider ConvNeXt-T~\citep{liu2022convnet} and ViT-S~\citep{dosovitskiy2020image} for completeness. Our pre-processing follows standard supervised training~\citep{liu2022convnet}. Specifically, during training, the classifiers process randomly augmented crops of $224 \times 224$ images. For validation, images are resized to 256 pixels on the shorter side, preserving the aspect ratio, and then center cropped to $224 \times 224$. For CIFAR-10 experiments, we initialize the classifier with ResNet-50 pre-trained on ImageNet and also provide the results trained from scratch (see Appendix~\ref{app:result_of_cifar}). For ImageNet, we train ResNet-50 from scratch.

\textbf{Diffusion model.} For CIFAR-10 generation, we consider two diffusion models: EDM~\citep{karras2022elucidating} and U-ViT~\citep{bao2023all}. For ImageNet-256, we explore three diffusion models: EDM2~\citep{karras2023analyzing}, U-ViT-H/2~\citep{bao2023all}, and DiT-XL/2~\citep{peebles2023scalable}. Since EDM2 is originally designed for ImageNet-512 generation, we resize its generated images from 512 to 256 resolution, with EDM2-XXL achieving an FID of 2.14, comparable to U-ViT-H/2 and DiT-XL/2. For ImageNet-512 generation, we use EDM2-XXL~\citep{karras2023analyzing}. For COCO generation, we consider three text-to-image diffusion models: Pixart-$\alpha$~\citep{chen2023pixart}, SDXL~\citep{podell2023sdxl}, and Playground-v2.5~\citep{li2024playground}.

\textbf{Evaluation.} We use the top-1 accuracy on the validation set to evaluate classification performance.

\textbf{Abbreviations and Notation.} In the figures and tables following this paper, the following abbreviations are used: CIFAR-10 (\emph{C}), ImageNet (\emph{I}), U-ViT (\emph{U}), EDM (\emph{E}), DiT-XL/2 (\emph{D}). Besides, "Distribution combinations" refer to multiple distributions, where the number of distributions corresponds to the number of classes in distribution classification.

\section{Generated distributions are easily classified as generated}
\label{sec:phenomenon1}
In this section, we explore how the distribution generated by diffusion models, known for their success in image generation, differs from the real distribution. To analyze this, we use a distribution classification task to assess the extent of the difference. Besides, we also extend our evaluation to other generative models, such as GANs, in Appendix~\ref{app:gan_label_to_image}.

\subsection{Classifiers cannot distinguish samples from the same distribution}
First, it is crucial to evaluate whether classifiers are indeed unable to distinguish samples from the same distribution. As shown in Tab.~\ref{tab:classifier_same_distribution}, we report the classification accuracy for the same distributions in the label-to-image scenario. Our findings indicate that classifiers are unable to distinguish samples from the same distribution, initially demonstrating the validity of the distribution classification task.

\subsection{Classifiers achieve high accuracy across various settings}
\begin{figure}[t]
    \centering
    \includegraphics[width=.7\linewidth]{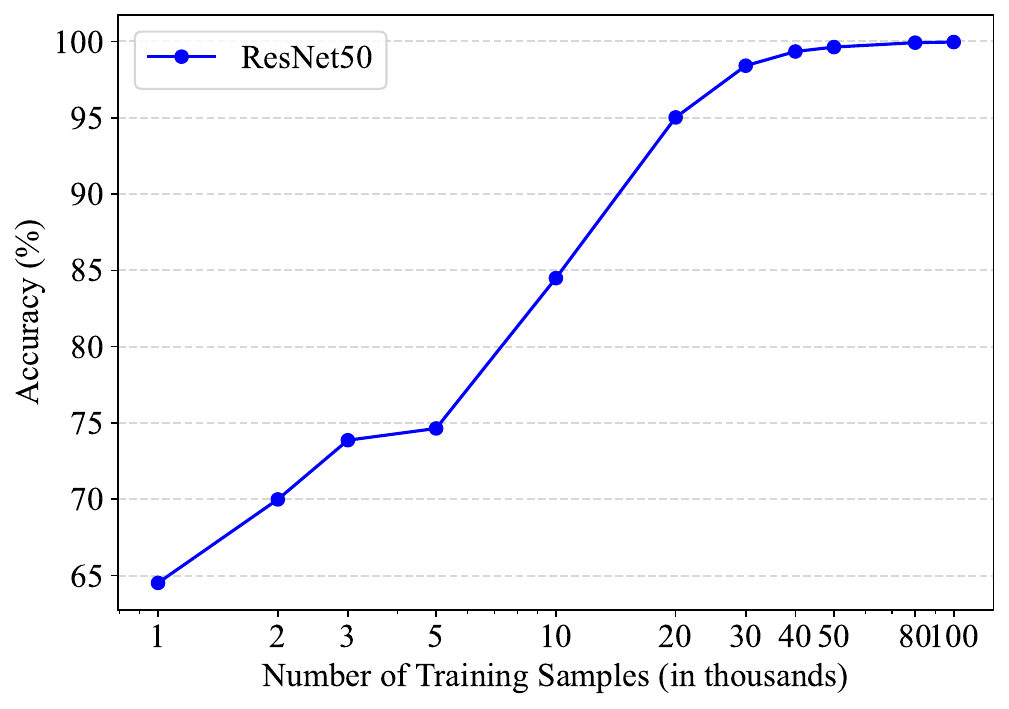}
    \caption{\textbf{Binary distribution classification on \emph{label-to-image}}. A positive correlation is observed between accuracy and the number of training samples. With only 50k samples per distribution, classifiers achieve over 99.5\% accuracy.}
    \label{fig:model_accuracy_vs_training_examples}
\end{figure}

\textbf{Label-to-image.} As shown in Tab.~\ref{tab:various_architecture_label_to_image}, we report the binary distribution classification accuracy for various combinations of real and generated distributions in the label-to-image scenario. Surprisingly, despite the low FID scores achieved by diffusion models, neural network-based classifiers still identify significant differences between the generated and real distributions. Specifically, all classifiers easily distinguish real from generated images across different dataset combinations, with most achieving over 98\% accuracy in distribution classification. Moreover, we observe a positive correlation between distribution classification accuracy and the number of training samples for the combination of U-ViT-H/2 and ImageNet-256 (see Fig.~\ref{fig:model_accuracy_vs_training_examples}). This suggests that as sample size increases, classifiers become more effective at distinguishing between distributions. Notably, with only 50k samples per distribution (less than 10\% of the ImageNet samples used to train the generator), classifiers achieve over 99.5\% accuracy. Furthermore, to investigate the impact of data diversity on distribution classification, we scale up the sample numbers per distribution. Using U-ViT-H/2, we generate 1.28 million images, combined with the full ImageNet training set, the classifiers achieve 100\% accuracy. Besides, we extend our evaluation to other generative models, including GANs and Flow Matching Models (see results in Appendix.~\ref{app:gan_label_to_image}).

\begin{table}[t]
    \small
    \centering
    \begin{tabular}{ccc}
    \toprule
        \textbf{Distribution combinations}  &  \textbf{Classifier} & \textbf{Accuracy (\%)} \\
    \midrule
        \multirow{3}{*}{$\text{\emph{U}}$~\citep{bao2023all}, U} &  ResNet-50 & 50.83 \\
        & ViT-S & 50.57 \\
        & ConvNeXt-T & 50.18 \\
    \midrule
        \multirow{3}{*}{$\text{\emph{E}}$~\citep{karras2022elucidating}, E}  &  ResNet-50 & 50.67 \\
        & ViT-S & 50.44 \\
        & ConvNeXt-T & 50.14 \\
    \bottomrule
    \end{tabular}
    \caption{\textbf{Binary distribution classification on \emph{the same distribution}}. Classifiers are indeed unable to distinguish samples from the same distribution, demonstrating the validity of our approach.}
    \label{tab:classifier_same_distribution}
    \vspace{-.15cm}
\end{table}

\begin{table}[t]
    \small
    \centering
    \begin{tabular}{ccccc}
    \toprule
        \textbf{Distribution combinations} &  \textbf{Classifier} & \textbf{Accuracy (\%)} \\
    \midrule
        \multirow{3}{*}{\emph{C}, $\text{\emph{U}}$~\citep{bao2023all}} &  ResNet-50 & 99.92 \\
        & ViT-S & 98.04 \\
        & ConvNeXt-T & 99.96 \\
    \midrule
        \multirow{3}{*}{\emph{C}, $\text{\emph{E}}$~\citep{karras2022elucidating}} 
        &  ResNet-50 & 96.25 \\
        & ViT-S & 89.38 \\
        & ConvNeXt-T & 98.43 \\
    \midrule
        \multirow{3}{*}{{\emph{I}-256}, {\emph{U}-H/2}~\citep{bao2023all}} &  ResNet-50 & 99.95 \\
        & ViT-S & 98.06 \\
        & ConvNeXt-T & 99.89\\
    \midrule 
        \multirow{1}{*}{{\emph{I}-256}, {\emph{U}-H/2}~\citep{bao2023all}}$^\dagger$ &  ResNet-50 & 100.00 \\
    \midrule
        \multirow{3}{*}{\emph{I}-512, {\emph{E}2-XXL}~\citep{karras2023analyzing}} &  ResNet-50 & 99.73\\
        & ViT-S & 95.15 \\
        & ConvNeXt-T & 86.44 \\
    \bottomrule
    \end{tabular}
    \caption{\textbf{Binary distribution classification on \emph{label-to-image}}. 
    All classifiers yield high accuracy on various combinations of datasets and generative models. FID values for each generative model are: \emph{U}: 3.11, \emph{E}: 1.79, \emph{U}-H/2: 2.29, and \emph{E}2-XXL: 1.81. $^\dagger$ Indicates experiments conducted on the full ImageNet dataset.}
    \label{tab:various_architecture_label_to_image}
\end{table}

\textbf{Text-to-image.} As shown in Tab.~\ref{tab:various_architecture_text_to_image}, we report the distribution classification accuracy in the text-to-image scenario. We consider combinations of four distributions: COCO~\citep{coco}, Pixart-$\alpha$~\citep{chen2023pixart}, SDXL~\citep{podell2023sdxl}, and Playground-v2.5~\citep{li2024playground}. 
In this task, classifiers are not only required to distinguish between real and generated distributions but also correctly identify the specific generated distribution an image comes from. Remarkably, classifiers achieve an accuracy of over 76\% across all settings. Furthermore, we observe that increasing the training sample size from 5k to 10k results in significant accuracy gains, consistent with the patterns seen in the label-to-image scenario.

\subsection{Self-supervised classifiers can also identify generated images}
In addition to the supervised learning protocol, we also explore self-supervised learning. Specifically, we use self-supervised pre-trained models, such as MAE~\citep{he2022masked} and MoCo v3~\citep{chen2021empirical}, to extract image features followed by training a linear classifier on these features, referred to as the self-supervised classifier. The distribution classification accuracy of these classifiers is in Tab.~\ref{tab:self_supervised_learning}. Notably, the self-supervised classifiers also demonstrate a strong ability to distinguish between real and generated distributions. Moreover, we observed two intriguing phenomena: First, unlike in semantic classification tasks on ImageNet, the masked image modeling-based MAE outperforms the contrastive learning-based MoCo v3 in identifying generated images related to ImageNet. Second, while self-supervised classifiers achieve relatively high accuracy on ImageNet-related tasks, their performance drops significantly when applied to CIFAR-10.
We hypothesize this decline can be contributed to two factors: (1) MAE and MoCo v3 are pre-trained on ImageNet, not CIFAR-10, and (2) diffusion models perform exceptionally well on CIFAR-10 image generation, making it challenging for a simple linear classifier to differentiate between real and generated images. Besides, we also evaluate self-supervised classifiers in text-to-image scenarios (see results in Appendix.~\ref{app:self_supervised_text_to_image}).


\begin{table}[t]
    \small
    \centering
    \begin{tabular}{ccc}
    \toprule
        \textbf{Training samples} & \textbf{Model} & \textbf{Accuracy (\%)} \\
    \midrule
         5k & ResNet-50 & 88.28 \\
         10k & ResNet-50 & 93.45 \\
         5k & ViT-S & 76.33 \\
         10k & ViT-S & 83.53 \\
         5k & ConvNeXt-T & 80.53 \\
         10k & ConvNeXt-T & 85.13 \\
    \bottomrule
    \end{tabular}
    \caption{\textbf{Four-way distribution classification on \emph{text-to-image}}. All classifiers yield high accuracy in distinguishing four distributions: COCO~\citep{coco}, Pixart-$\alpha$~\citep{chen2023pixart}, SDXL~\citep{podell2023sdxl}, and Playground-v2.5~\citep{li2024playground}, using only 5k or 10k training samples per dataset. Notably, as the number of training samples increases, the accuracy of distribution classification consistently improves.}
    \label{tab:various_architecture_text_to_image}
\end{table}

\begin{table}[t]
    \small
    \centering
    \begin{tabular}{cccc}
    \toprule
        \textbf{Distribution combinations} & \textbf{SSL method} & \textbf{Accuracy} \\ 
    \midrule
        \multirow{2}{*}{\emph{C}, \emph{U}~\citep{bao2023all}} & MAE & 71.83 \\ 
        & MoCo v3 & 73.39 \\
    \midrule    
        \multirow{2}{*}{\emph{C}, \emph{E}~\citep{karras2022elucidating}} & MAE & 64.83 \\ 
         & MoCo v3 & 67.69 \\ 
    \midrule
        \multirow{2}{*}{\emph{I}-256, \emph{U}-H/2~\citep{bao2023all}} & MAE & 91.65 \\ 
        & MoCo v3 & 77.38 \\ 
    \midrule
        \multirow{2}{*}{\emph{I}-512, \emph{E}2-XXL~\citep{karras2023analyzing}} & MAE & 81.80 \\ 
        & MoCo v3 & 74.30 \\ 
    \bottomrule
    \end{tabular}
    \caption{\textbf{Self-supervised classifiers demonstrate the ability to distinguish between real and generated distributions.} For each self-supervised method, we use ViT-B as the backbone.}
    \label{tab:self_supervised_learning}
\end{table}

\section{Classifier contradictions with FID and human judgments}
\label{sec:phenomenon2}
Previously, we demonstrated that in most settings, classifiers can effortlessly distinguish between real and generated distributions, indicating that the generated distribution is still far from the real one. In this section, we explore the intriguing contradictions between classifiers and widely used evaluation protocols, such as FID and human judgments, through two sets of experiments. The first experiment compares EDM2-XS and EDM2-XXL, which share the same network architectures and training methodologies but differ in the number of parameters, resulting in FID scores of 2.91 and 1.81, respectively. The second experiment examines DiT-XL, U-ViT-H/2, and EDM2-XXL, which, despite having similar FIDs, vary in network architecture, sampling methods, and training protocols.

\begin{figure}[t]
    \centering
    \includegraphics[width=.8\linewidth]{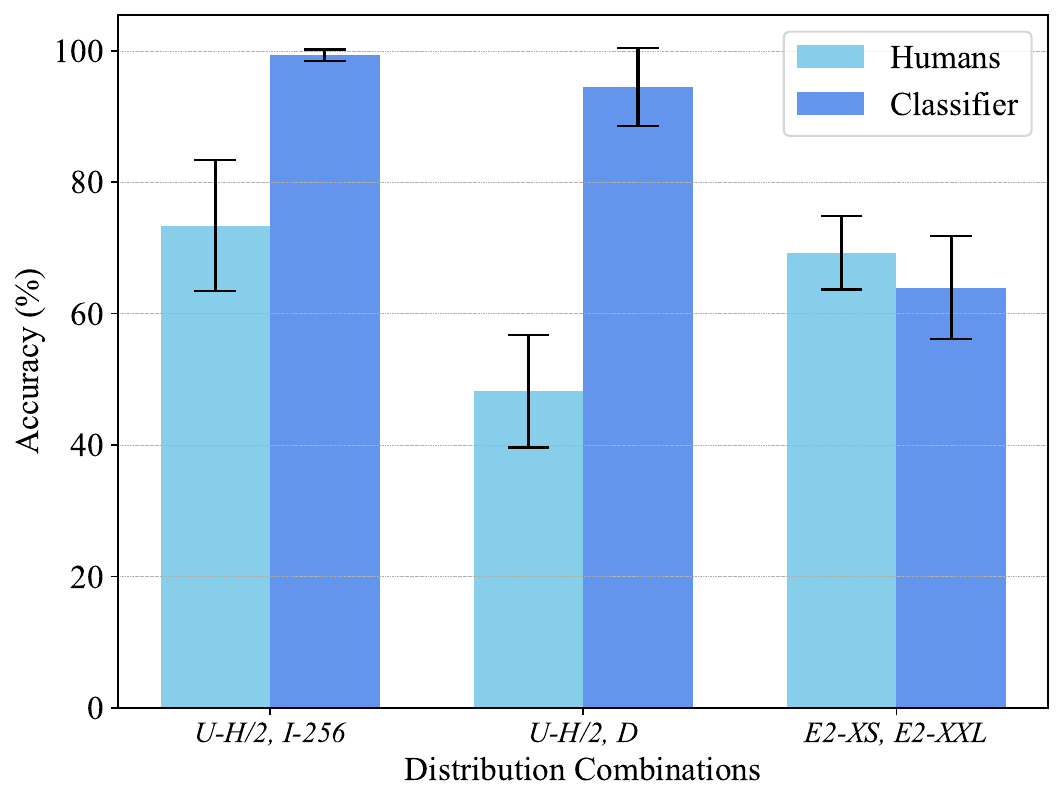}
    \caption{\textbf{User study.} Classifiers can easily distinguish between diffusion models with similar FIDs (U-H/2, D) but struggle with models from the same family that differ in parameters (E2-XS, E2-XXL). In contrast, humans show the opposite trend.}
    \label{fig:user_study}
\end{figure}

To analyze human judgments, we conducted a user study to evaluate the human ability to distinguish between different distributions (see the user interface in Fig.~\ref{fig:user_study_appendix}). Further details of the study are provided in Appendix~\ref{app:user_study}. The results are shown in Fig.~\ref{fig:user_study}, where individual variability is represented using error bars, indicating the standard deviation across participants/classifier performance.

\begin{table}[t]
    \small
    \centering
    \begin{tabular}{cccc}
    \toprule
        \textbf{Combinations} & \textbf{Classifier} & \textbf{Accuracy (\%)} \\
    \midrule
        \multirow{3}{*}{\emph{E}2-XS, \emph{E}2-XXL} & ResNet-50 & 74.91\\
          & ConvNeXt-T & 57.35 \\
          & ViT-S & 59.57 \\
    \bottomrule
    \end{tabular}
    \caption{\textbf{Classifiers struggle to differentiate between the smallest and largest EDM2 models.} Despite notable differences in FID (E2-XS: 2.91, E2-XXL:1.81), the classifiers show limited ability to accurately distinguish between these EDM2 variants.}
    \label{tab:edm2_compare}
\end{table}

In our first set of experiments, we analyze models within the same family: EDM2-XS and EDM2-XXL, the smallest and largest models in the EDM2 family, respectively. As shown in Tab.~\ref{tab:edm2_compare}, we report the distribution classification accuracy for distinguishing between EDM2-XS and EDM2-XXL. The results show that classifiers struggle to differentiate between models from the same family. Only ResNet-50 shows some ability to distinguish them, while the other classifiers perform close to random chance. This contrasts sharply with the task of distinguishing between ImageNet-512 and EDM2-XXL (in Tab.~\ref{tab:various_architecture_label_to_image}), where all classifiers achieve over 86\% accuracy. Thus, compared to distinguishing between diffusion models and the real distribution, neural networks view models within the same family—due to their shared inductive biases—as learning similar distributions, making them difficult to distinguish. Furthermore, in our user study (see Fig.~\ref{fig:user_study}), unlike the other tasks where classifiers significantly outperform humans, in the task of distinguishing between models within the same family, humans perform comparably to the classifiers. This observation highlights a limitation of neural network-based classifiers: despite the noticeable difference in FID scores between EDM2-XS and EDM2-XXL (2.91 and 1.81, respectively), the classifiers struggle to differentiate between them.

\begin{table}[t]
    \small
    \centering
    \begin{tabular}{ccc}
    \toprule
        \textbf{Distribution combinations} & \textbf{Model} & \textbf{Accuracy} \\
    \midrule
        \multirow{3}{*}{\emph{U}-H/2, \emph{D}} & ResNet-50 & 99.66 \\
         & ConvNeXt-T & 97.59 \\
         & ViT-S & 86.13 \\
    \midrule
        \multirow{3}{*}{\emph{U}-H/2, \emph{D}, \emph{I}-256} & ResNet-50 & 99.87 \\
         & ConvNeXt-T & 99.77 \\
         & ViT-S & 95.90 \\
    \midrule
        \multirow{3}{*}{\emph{U}-H/2, \emph{D}, \emph{I}-256, \emph{E}2-XXL} & ResNet-50 & 99.91 \\
         & ConvNeXt-T & 99.92 \\
         & ViT-S & 98.13 \\
    \bottomrule
    \end{tabular}
    \caption{\textbf{Distribution classification accuracy for various combinations of diffusion models with similar FIDs and ImageNet}. Notably, accuracy improves as more distributions are added.}
    \label{tab:similar_fid}
\end{table}

In our second group of experiments, we analyze different diffusion models with similar FIDs. As shown in Tab.~\ref{tab:similar_fid}, we present the distribution classification accuracy for three combinations of diffusion models with similar FIDs. We observe that classifiers can easily distinguish between these models, achieving extremely high accuracy, while humans perform close to random guessing (see Fig.~\ref{fig:user_study}). This highlights a key advantage of classifiers over widely used evaluation protocols: despite the FID scores being very close and humans being unable to differentiate between the models, classifiers can easily identify the differences. Moreover, an intriguing phenomenon occurs when additional distributions are introduced: the classifier's accuracy improves further. This contrasts with the findings in~\citet{liu2024decade}. A possible explanation is that all of these distributions are either directly related to ImageNet or generated by diffusion models trained on it. As new distributions are added, the classifier can leverage newly learned patterns to enhance the classification accuracy for previous combinations.

\section{Implications of distribution classification task}
Previously, we demonstrated the classifier's ability to distinguish real from generated distributions and the contradictions between classifiers and evaluation methods like FID and human judgments. In this section, we show some important implications of the distribution classification task.

\subsection{Diagnosing problem within diffusion models}
\label{sec:evaluate_alignment}
A natural implication of classifiers is to diagnose the problem within diffusion models. Our approach complements traditional metrics like FID and human judgments by offering a more challenging evaluation method. It can also be extended to analyze specific features of generated data, providing deeper insights into the intrinsic characteristics of diffusion models, such as spatial and frequency features. In this section, we focus on advanced diffusion transformers, specifically U-ViT-H/2~\citep{bao2023all}, and investigate two key aspects: spatial features and frequency ranges. Our objective is to identify which spatial features diffusion models learn well, making them harder to distinguish, and which they struggle with, making them easier to detect. Additionally, we assess performance across different frequency ranges to pinpoint where diffusion models succeed and where they fail. For this analysis, we use U-ViT-H/2 trained on ImageNet-256, with ResNet-50 as the classifier.

\begin{table}[!t]
    \small
    \centering
    \begin{tabular}{cccc}
    \toprule
        \textbf{Combinations} & \textbf{Crop size} & \textbf{Accuracy (\%)} \\
    \midrule
        \multirow{7}{*}{\emph{U}-H/2, \emph{I}-256} & None & 99.96 \\
         & 128$^\dagger$ & 99.96 \\
         & 64$^\dagger$ & 99.97 \\
         & 32$^\dagger$ & 99.97 \\
         & 16$^\dagger$ & 99.78 \\
         & 128$^\star$ & 99.97 \\
         & 64$^\star$ & 99.97 \\
    \bottomrule
    \end{tabular}
    \caption{\textbf{Classifiers achieve high accuracy across crop sizes}. $^\dagger$ denotes center crop. $^\star$ denotes random crop, where results are averaged over four trials.}
    \label{tab:various_crop_size_imagenet}
\end{table}

\begin{figure*}[t]
    \centering
    \includegraphics[width=.8\linewidth]{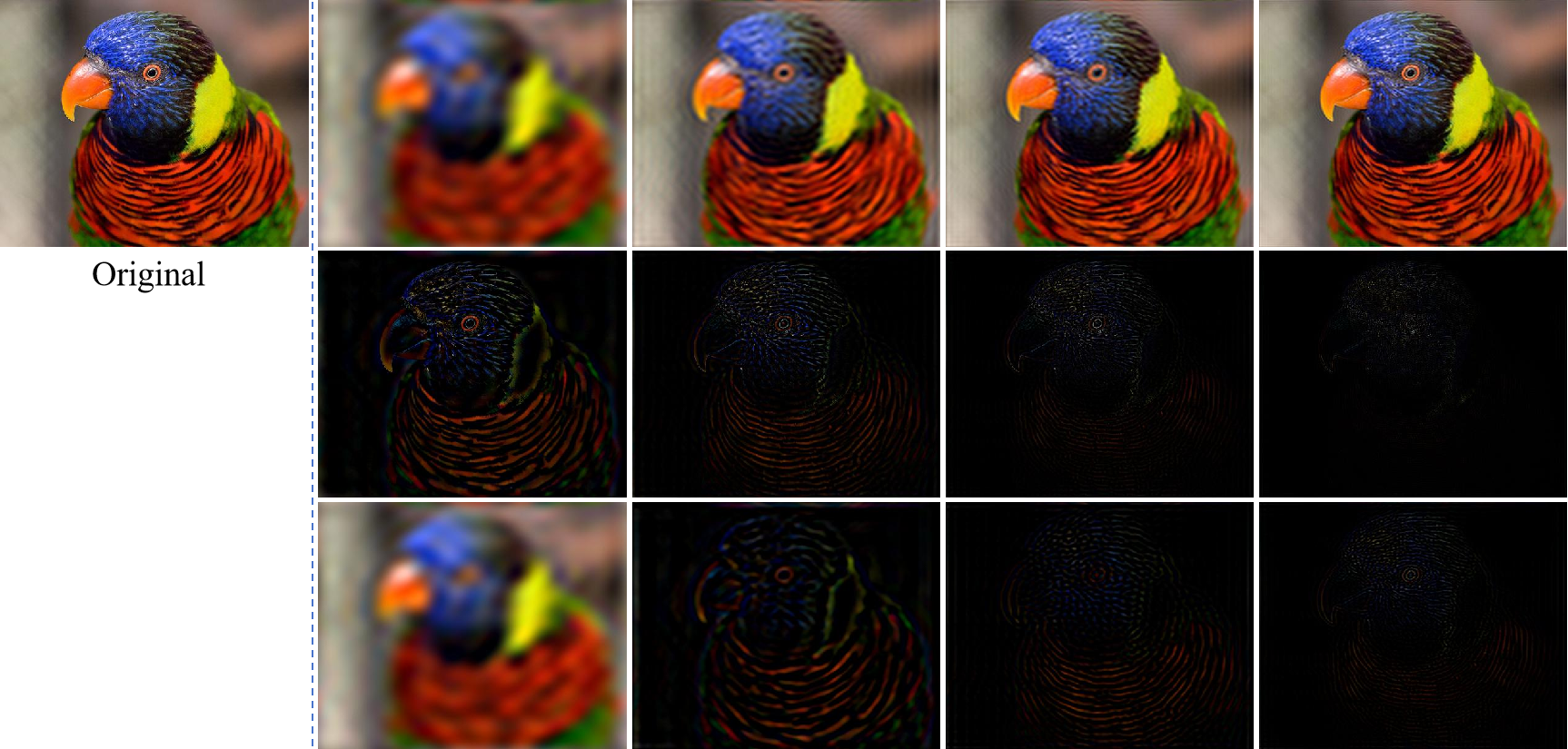}
    \caption{\textbf{Visualization of frequency domain processing.} \emph{Top}: Low-pass filters, \emph{Middle}: High-pass filters, each processed with increasing thresholds: 10, 30, 50, 100. \emph{Bottom}: Band-pass filters, processed with band thresholds: 0-10, 10-30, 30-50, 50-100.}
    \label{fig:frequency_process}
\end{figure*}

\begin{figure*}[t]
    \centering
    \begin{subfigure}[b]{0.325\linewidth}
        \includegraphics[width=\linewidth]{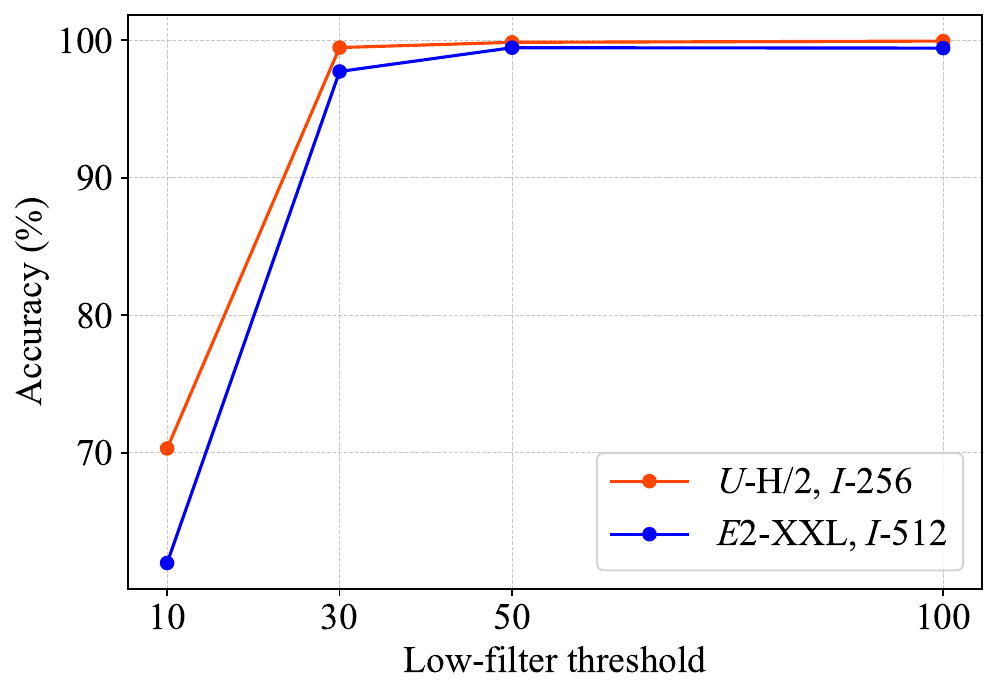}
        \caption{Accuracy vs. low-frequency filter threshold}
        \label{fig:resnet50_low_filter_experiment}
    \end{subfigure}
    \hfill
    \begin{subfigure}[b]{0.325\linewidth}
        \includegraphics[width=\linewidth]{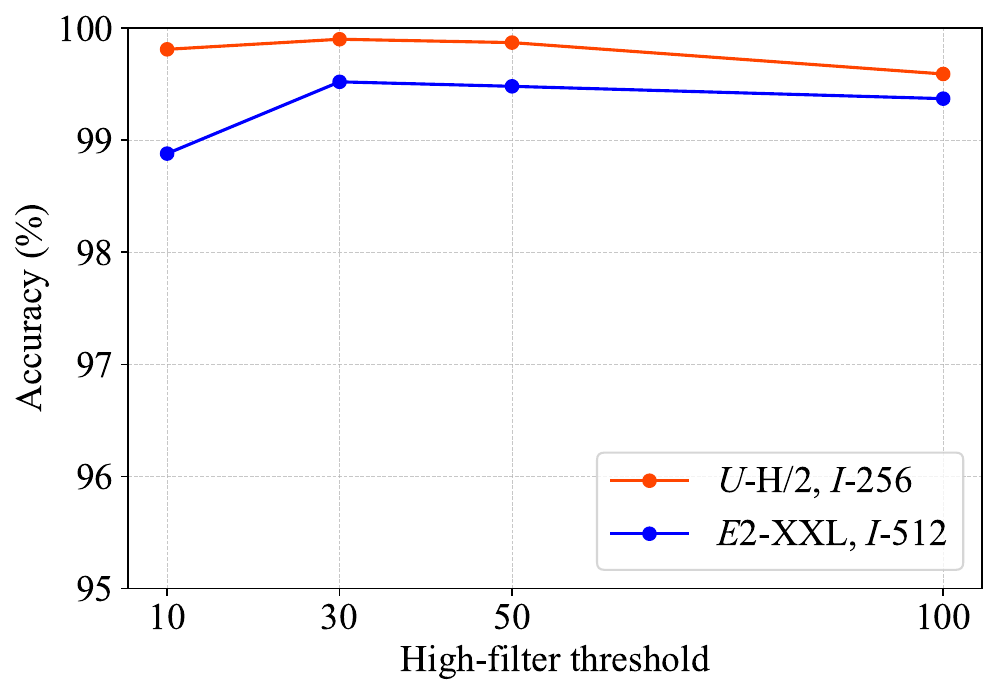}
        \caption{Accuracy vs. high-frequency filter threshold}        \label{fig:resnet50_high_filter_experiment}
    \end{subfigure}
    \hfill
    \begin{subfigure}[b]{0.325\linewidth}
        \includegraphics[width=\linewidth]{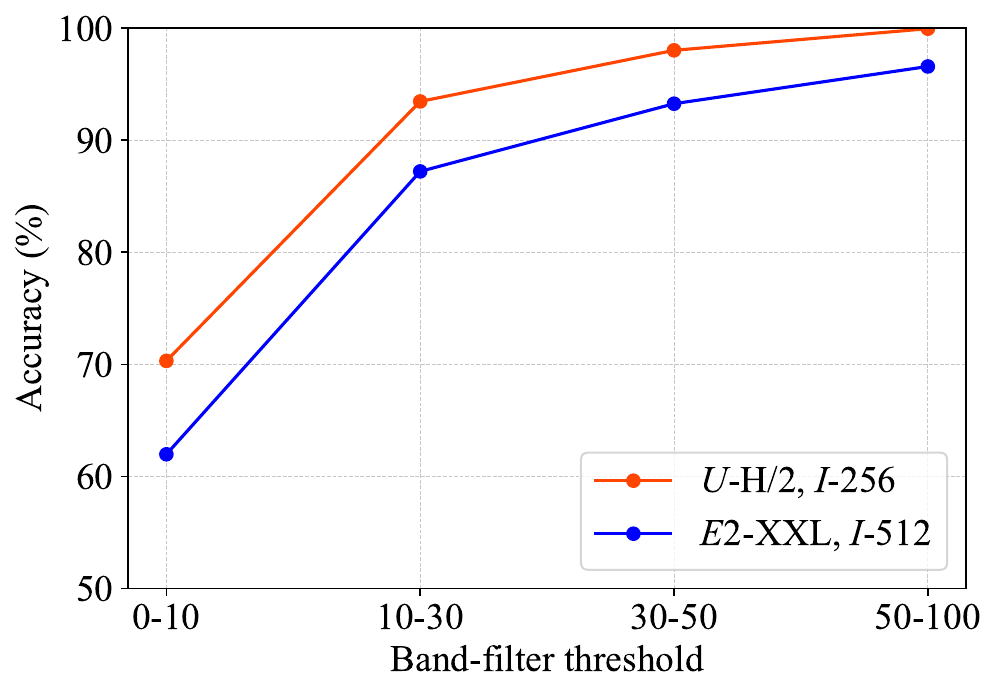}
        \caption{Accuracy vs. band-frequency filter threshold}        \label{fig:resnet50_band_filter_experiment}
    \end{subfigure}
    \caption{\textbf{Classifier achieves high accuracy across various combinations with limited frequency components,} except when only minimal low-frequency components are present (in subfig (a) low-pass filter threshold = 10, in subfig (c) band-pass filter threshold: 0-10).}
    \label{fig:resnet50_filter_experiment}
\end{figure*}

In our first set of experiments, we focus on the spatial features. We start by progressively reducing central information through center-cropping images from a resolution of 256 down to 128, 64, 32, and 16, followed by training the classifier from scratch. The results, shown in Tab.~\ref{tab:various_crop_size_imagenet}, indicate that even with minimal central information, the classifier can accurately distinguish between real and generated distributions. We then assess whether classifiers can differentiate based on any arbitrary part of the image. We cropped images at resolutions of 128 and 64, and randomly selected four sections per image for training (see Fig.~\ref{fig:crop_visualization} in Appendix for visualizations). The classifiers achieved accuracies of $99.97 \pm 0.004$\% for 128 and $99.97 \pm 0.007$\% for 64 (see Tab.~\ref{tab:various_crop_size_imagenet}). These results demonstrate that even with the advanced diffusion transformer U-ViT-H/2, which achieves an FID of 2.29, classifiers can reliably distinguish between real and generated images using any spatial part of the image.

In our second set of experiments, we focus on the frequency aspect. We preprocess images using low-pass, high-pass, and band-pass filters as described in Sec.~\ref{sec:frequency_filter}. For the low-pass and high-pass filters, we apply thresholds of 10, 30, 50, and 100, while for the band-pass filter, we use frequency intervals of 0-10, 10-30, 30-50, and 50-100. Fig.~\ref{fig:frequency_process} provides examples of the original and filtered images. As shown in Fig.~\ref{fig:resnet50_filter_experiment}, using ResNet-50 as the classifier, we observe that classifiers struggle to maintain high accuracy when low-frequency information is minimal but perform well across all other frequency intervals, even with only high-frequency components (e.g., threshold of 100). These results suggest that diffusion models effectively learn low-frequency features, making it more difficult for classifiers to distinguish real from generated images in this range. However, in higher frequency bands, diffusion models perform worse, making it easier for classifiers to identify generated images.

\subsection{Evaluating the use of generated data}
Another natural implication is using classifiers to evaluate the use of generated data. Our previous findings show a significant difference between generated and real distributions, raising the question of whether this mismatch causes undesirable behaviors in models trained on synthetic data. One well-known phenomenon is that training generative models on data produced by previous generations in a continuous loop leads to a gradual decline in output quality and diversity~\citep{shumailov2024ai, alemohammad2024self}, ultimately resulting in model collapse, known as Model Autophagy Disorder (MAD). This is surprising, as data generated by modern diffusion models often exhibit high realism. However, using this data to train next-generation models triggers MAD. Our approach provides insight into this issue: although these models generate realistic images, the distributions they produce remain distinct from the real distribution—especially when viewed by classifiers—leading to a distribution mismatch. This mismatch accumulates, causing the generated distribution to drift. \citet{alemohammad2024self} address this by incorporating real samples into each generation's training data.

\begin{figure}[t]
    \centering
    \includegraphics[width=.8\linewidth]{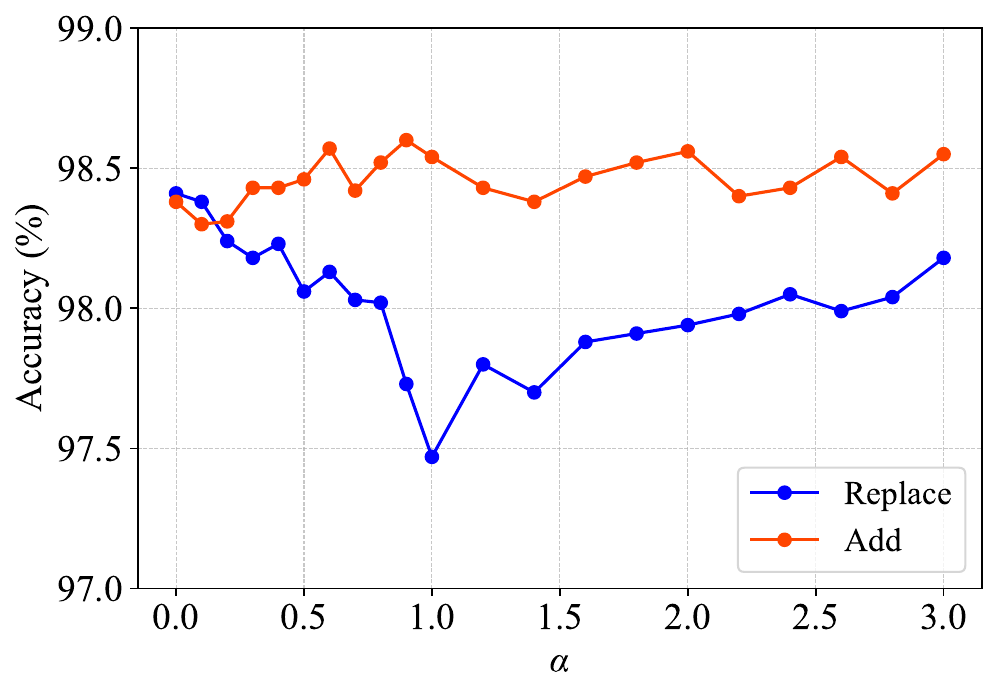}
    \caption{\textbf{Top-1 accuracy on the CIFAR-10 test set: Replacement vs. Augmentation.} $\alpha$ represents the ratio of EDM\_S mixed with CIFAR-10.}
    \label{fig:mix_exp_cifar10}
\end{figure}

To better understand this phenomenon, we further explore the impact of using generative model data in downstream tasks. Specifically, we investigate whether the distribution mismatch between real and generated data causes undesirable behaviors when replacing real data with generated data in tasks such as classification. To analyze this, we use EDM~\citep{karras2022elucidating} to generate the same number of samples as CIFAR-10, creating a synthetic dataset, EDM\_S, and perform both supervised and semi-supervised classification to assess the impact. For supervised classification, we train the classifier with ConvNeXt settings~\citep{liu2022convnet} for CIFAR-10, using the original test set for evaluation. To create the training dataset, we mix EDM\_S and CIFAR-10 with a ratio $\alpha$. We employ two strategies: one replaces $\alpha$ of CIFAR-10 with EDM\_S, where $\alpha$ represents the replacement ratio, and the other augments EDM\_S to CIFAR-10. As shown in Fig.~\ref{fig:mix_exp_cifar10}, we observe that due to the distribution mismatch, simply replacing samples leads to a decline in accuracy. As $\alpha$ increases from 0 to 1, the top-1 accuracy gradually declines. In contrast, when augmenting EDM\_S to CIFAR-10, accuracy generally improves for most $\alpha$ values. Additionally, as $\alpha$ increases from 1 to 3, the replacement strategy exhibits modest improvement in top-1 accuracy but remains consistently inferior to using only CIFAR-10. Conversely, the augmentation strategy maintains robust performance throughout this range. For semi-supervised classification, we use FreeMatch~\citep{wang2023freematch} and FlexMatch~\citep{zhang2021flexmatch}. As shown in Tab.~\ref{tab:semi_classification_cifar10}, replacing CIFAR-10 with EDM\_S degrades semi-supervised learning performance for both FlexMatch and FreeMatch, consistent with supervised findings. However, when augmenting real data with generated data through DPT~\citep{you2024diffusion}, we observe improved model performance compared to using real data alone.

The distribution mismatch between generated and real data causes issues like "MAD" or lower performance when generated data replaces real data. Augmenting real data with generated data, however, is more beneficial.

\begin{table}[t]
    \small
    \centering
    \begin{tabular}{lcc}
    \toprule
    {Method}  & \multicolumn{2}{c}{Error rate $\downarrow$ } \\
    given $\#$ labels per class (label fraction) &  4 (0.08\%) \\
    \midrule
    FlexMatch$^\star$\citep{zhang2021flexmatch} & 4.97{\scriptsize $\pm$0.06} \\
    FlexMatch$^\dagger$ & 5.85{\scriptsize $\pm$0.02} \\
    \midrule
    FreeMatch$^\star$\citep{wang2023freematch} & 4.90{\scriptsize $\pm$0.04} \\
    FreeMatch$^\dagger$ & 5.94{\scriptsize $\pm$0.10} \\
    FreeMatch$^\ddagger$\citep{you2024diffusion}& 4.68{\scriptsize $\pm$0.17} \\
    \bottomrule
    \end{tabular}
    \caption{\textbf{Augmenting real data with generated data is more effective than replacing it.} $^\star$ indicates results from training with CIFAR-10 (real data); $^\dagger$ and $^\ddagger$ indicate results where CIFAR-10 is replaced or augmented with EDM\_S, respectively.}
    \label{tab:semi_classification_cifar10}
\end{table}

\subsection{Enhancing Realism in Generated Images}
Our previous experiments demonstrated that classifiers can readily distinguish generated images from real ones. This naturally leads to the question of how to enhance the realism of generative models. Motivated by the classifier guidance technique in ADM~\citep{dhariwal2021diffusion}, which improves the alignment of generated images with specified labels, we investigate a similar approach for improving realism. We conduct an experiment where we first generate 100,000 synthetic images using ADM. These are combined with 100,000 ImageNet images to form a training dataset.  A ResNet-50 classifier is then trained on noisy versions of these images, where the noise is introduced by randomly sampling timesteps (1-1000) and applying the corresponding forward diffusion process. The classifier is trained to distinguish between real and generated images. Finally, we utilize the trained classifier's guidance to steer the diffusion model towards generating images that are more aligned with the real image distribution.  Our results show that classifiers identify 96.19\% of the original (unguided) generated images as fake. However, with classifier guidance, this percentage decreases to 79.88\%. This experiment demonstrates that classifier guidance can enhance the realism of images produced by diffusion models.
\section{Conclusion}
\label{sec:conclusion}
Our paper proposes the distribution classification tasks to explore a fundamental question: \textit{How far is the generated distribution from the real distribution?} Our findings show that classifiers can effectively distinguish between diffusion-generated and real distributions, even when diffusion models achieve low FID scores and generate lifelike images. Additionally, our study shows contradictions between classifier performance and evaluation metrics such as FID and human judgments. Finally, our methodology offers several key implications: it serves as a natural diagnostic tool for diffusion models, provides a reasonable explanation for Model Autophagy Disorder (MAD), guides the effective use of generated data, and enhances the realism of generated images through classifier guidance.

\textbf{Limitations}. Our findings are based on specific datasets and models, potentially limiting generalizability. Future work can explore broader datasets and models for validation.

\section*{Acknowledgement}
This work was sponsered by the Beijing Nova Program(No. 20220484044); National Natural Science Foundation of China (No. 92470118); Beijing Natural Science Foundation (No. L247030); Major Innovation \& Planning Interdisciplinary Platform for the ``Double-First Class" Initiative, Renmin University of China; the Fundamental Research Funds for the Central Universities, and the Research Funds of Renmin University of China (22XNKJ13). The work was partially done at the Engineering Research Center of Next-Generation Intelligent Search and Recommendation, Ministry of Education.

{
    \small
    \bibliographystyle{ieeenat_fullname}
    \bibliography{main}
}

\clearpage
\appendix
\setcounter{page}{1}
\maketitlesupplementary

\section{Background}
\subsection{Diffusion model}
Diffusion models~\citep{sohl2015deep, song2020score, ho2020denoising} gradually inject noise into data $\boldsymbol{x}$ during the forward process:
\begin{align}
    \boldsymbol{z}_t = \sqrt{\bar{\alpha}_t} \boldsymbol{x} + \sqrt{1 - \bar{\alpha}_t} \boldsymbol{\epsilon},
\end{align}
and remove noise to generate data in the reverse process. Diffusion models typically use a noise prediction network $\boldsymbol{\epsilon}_{\boldsymbol{\theta}}(\boldsymbol{z}_t, t)$ to predict the noise $\boldsymbol{\epsilon}$ added to $\boldsymbol{z}_t$. Noise prediction loss is defined as:
\begin{align}
\label{eq:loss_diffusion}
\mathcal{L} = \mathbb{E}_{t, \boldsymbol{x}_0, \boldsymbol{\epsilon}} [\|\boldsymbol{\epsilon}_{\boldsymbol{\theta}}(\boldsymbol{z}_t, t) - \boldsymbol{\epsilon}\|_2^2].
\end{align}

\subsection{Frequency filter}
\label{sec:frequency_filter}
The frequency content of an image represents the rate of pixel value changes. Low-frequency components capture overall shapes and gradual grayscale changes, while high-frequency components reflect fine details like edges and textures. Notably, convolutional neural networks can detect high-frequencies that are often imperceptible to humans~\citep{wang2020high}. In our paper, we implement a rectangular filter and discuss the choice of this filter in Appendix~\ref{app:choice_filter}.

\textbf{Low-pass filters} are implemented as follows: Compute the Fourier transform of the image \( f(x, y) \) using \( F(u, v) = \text{FFT}(f(x, y)) \), and then center the spectrum with \( F_c(u, v) = \text{fftshift}(F(u, v)) \). Define a rectangular mask \( H(u, v) \) in frequency domain, where $M$ and $N$ are the image dimensions:
\begin{align}
H(u, v) = 
\begin{cases} 
1 & \text{if } \left| u - \frac{M}{2} \right| \leq \text{threshold} \\ 
  & \quad \text{and } \left| v - \frac{N}{2} \right| \leq \text{threshold}, \\
0 & \text{otherwise}.
\end{cases}   
\end{align}
Apply the mask by multiplying \( G(u, v) = F_c(u, v) \odot H(u, v) \), reverse the shift with \( G_c(u, v) = \text{ifftshift}(G(u, v)) \), and transform back to the spatial domain using \( g(x, y) = \text{IFFT}(G_c(u, v)) \).

\textbf{High-pass filters} are implemented similarly, but with the mask defined to pass high frequencies:
\begin{align}
H(u, v) = 
\begin{cases} 
0 & \text{if } \left| u - \frac{M}{2} \right| \leq \text{threshold} \\ 
  & \quad \text{and } \left| v - \frac{N}{2} \right| \leq \text{threshold}, \\
1 & \text{otherwise}.
\end{cases}   
\end{align}

\textbf{Band-pass filters} combine low-pass and high-pass filters, allowing frequencies between low and high thresholds to pass while setting others to 0.

\section{Detail experiment settings}
\label{app:exp_settings}
We implement our experiments upon the official code of ConvNeXt~\citep{liu2022convnet}, U-ViT~\citep{bao2023all}, DiT~\citep{peebles2023scalable}, EDM~\citep{karras2022elucidating}, EDM2~\citep{karras2023analyzing}. We also utilize the official models of Pixart-$\alpha$~\citep{chen2023pixart}, SDXL~\citep{podell2023sdxl}, and Playground-v2.5~\citep{li2024playground}. The respective links and licenses are detailed in Tab.~\ref{tab:code_used_and_license}.

\begin{table*}[t!]
\vskip 0.1in
\small
\centering
\begin{tabular}{lcc}
\toprule
\textbf{Method} & \textbf{Link}  &  \textbf{License} \\
\midrule
ConvNeXt & \url{https://github.com/facebookresearch/ConvNeXt} &  MIT License \\
U-ViT & \url{https://github.com/baofff/U-ViT} &  MIT License \\
DiT & \url{https://github.com/facebookresearch/DiT} & CC BY-NC 4.0 \\
EDM & \url{https://github.com/NVlabs/edm} & CC BY-NC-SA 4.0 \\
EDM2 & \url{https://github.com/NVlabs/edm2} & CC BY-NC-SA 4.0 \\
\midrule
\textbf{Model} & \textbf{Link (add `https://huggingface.co/')}   &  \textbf{License} \\ 
\midrule
PixArt-$\alpha$ & \url{PixArt-alpha/PixArt-XL-2-1024-MS} & Open RAIL++-M \\ 
SDXL & \url{stabilityai/stable-diffusion-xl-base-1.0} & Open RAIL++-M \\
Playground-v2.5 & \url{playgroundai/playground-v2.5-1024px-aesthetic} & Playground v2.5 \\
\bottomrule
\end{tabular}
\caption{\label{tab:code_used_and_license} \textbf{Code links and licenses.}}
\end{table*}

We present the main experiment settings as follows. 

\textbf{Dataset.} For label-to-image, we consider the CIFAR~\citep{cifar} and ImageNet~\citep{imagenet} datasets, which are well-established and widely recognized benchmarks in the field of image generation. For text-to-image, we utilize the COCO2014 dataset~\citep{coco}, known for its rich annotations and diverse image content. For CIFAR-10, we use the real CIFAR-10 training set and validation set to construct the real distribution and use the diffusion model to generate 50k training images and 10k validation images to construct the generated distribution. In the case of ImageNet, we consider 256 and 512 resolutions, which are common resolutions for ImageNet image generation tasks. We randomly sample 100k training images and 50k validation images from the training set and the validation set of ImageNet to construct the real distribution and use the diffusion models to generate 100k training images and 50k validation images to construct the generated distribution. For the real dataset, we adopt the data processing method from ADM~\citep{dhariwal2021diffusion} to modify the ImageNet dataset into two common resolutions: ImageNet-256 and ImageNet-512, where the numbers indicate the resolution of the data. In addition, for the COCO2014 dataset, by default, we construct the real distribution by randomly sampling 10k training images and 1k validation images from the respective training and validation sets of COCO2014. Each image in this dataset is associated with five captions. To create the generated distribution, we randomly select one of the five captions for each real image to serve as a prompt. These prompts are then used by the diffusion models to generate an equivalent number of images to construct the generated distribution.

\textbf{Classifier.} By default, we employ the ResNet-50~\citep{he2016deep} as the classifier architecture. For completeness, we also consider ConvNeXt-T~\citep{liu2022convnet} and ViT-S~\citep{dosovitskiy2020image}. Our pre-processing protocol follows the standard supervised training approach~\citep{liu2022convnet}. Specifically, during training, classifiers process randomly augmented crops of $224 \times 224$ images. During validation, images are resized so that their smaller dimension reaches 256 pixels while preserving the original aspect ratio. Subsequently, these images are center cropped to $224 \times 224$ pixels before being fed into the model. For the experiments on CIFAR-10, we initialize our classifier with the ResNet-50, pre-trained on ImageNet. For completeness, we also present the result trained from scratch (see experiments in Appendix.~\ref{app:result_of_cifar}). In the case of ImageNet, we opt to train the ResNet-50 model from scratch.

\textbf{Diffusion model.} 
For the generation of CIFAR-10, we consider two diffusion models: EDM~\citep{karras2022elucidating} and U-ViT~\citep{bao2023all}. Both models have demonstrated strong generation performance on the CIFAR-10~\citep{cifar}. Quantitatively, EDM achieves an FID of 1.79, and U-ViT achieves an FID of 3.11. To improve efficiency, we modified U-ViT's sampling method from Euler-Maruyama to DPM-Solver~\citep{lu2022dpm} and reduced the sampling steps from 1,000 to 50. These adjustments resulted in U-ViT achieving an FID of 3.65 on CIFAR-10. For the generation of ImageNet-256, we explore three diffusion models: EDM2~\citep{karras2023analyzing}, U-ViT-H/2~\citep{bao2023all}, DiT-XL~\citep{peebles2023scalable}. Both DiT and U-ViT are prominent diffusion transformer architectures known for their scalability and strong performance. U-ViT-H/2 achieves an FID of 2.29 on ImageNet-256, and DiT-XL/2 achieves an FID of 2.27. We consider EDM2 to incorporate a UNet-based architecture, which was traditionally used before the rise of diffusion transformers. Since EDM2 is originally designed for ImageNet-512 generation, we resize the generated images from 512 to 256 resolution to suit our ImageNet-256 task. In this way, EDM2-XXL achieves an FID of 2.14 on this task, which is similar to the FID achieved by U-ViT and DiT. For the generation of ImageNet-512, we use EDM2~\citep{karras2023analyzing}, which achieves state-of-the-art performance on this task with an FID of 1.81. For the generation of COCO, we consider three state-of-the-art text-to-image diffusion models: Pixart-$\alpha$~\citep{chen2023pixart}, SDXL~\citep{podell2023sdxl}, and Playground-v2.5~\citep{li2024playground}. 

\textbf{Evaluation.} We use the top-1 accuracy on the validation set to evaluate classification performance.

\textbf{Training settings.} The complete training settings of ResNet-50 are reported in Tab.~\ref{tab:training-settings_cifar10} for combinations related to CIFAR-10 and Tab.~\ref{tab:training—settings_imagenet} for combinations related to ImageNet.

\begin{figure*}[t]
    \begin{minipage}{0.49\linewidth}
    \small
    \centering
    \begin{tabular}{@{}c|c@{}}
    \toprule
       \textbf{Config} & \textbf{Value} \\
    \midrule
        Optimizer & AdamW\\
        Learning rate & 4e-4\\
        Weight decay & 0.05 \\
        Optimizer momentum & $\beta_1, \beta_2$=0.9, 0.999\\
        Batch size & 256 \\
        Learning rate schedule & Cosine decay\\
        Warmup epochs & 0 \\
        Training epochs & 50 \\
        Augmentation & RandAug (9, 0.5)\\ 
        Label smoothing & 0.1 \\
        Mixup & 0 \\
        Cutmix & 0 \\
    \bottomrule
    \end{tabular}
    \captionof{table}{\textbf{Training settings for CIFAR-10.}}  
    \label{tab:training-settings_cifar10}
    \end{minipage}
    \hspace{0.00\linewidth}
    \begin{minipage}{0.49\linewidth}
    \small
    \centering
    \begin{tabular}{@{}c|c@{}}
    \toprule
       \textbf{Config} & \textbf{Value} \\
    \midrule
        Optimizer & AdamW \\
        Learning rate & 1e-3 \\
        Weight decay & 0.3 \\
        Optimizer momentum & $\beta_1, \beta_2$=0.9, 0.95\\
        Batch size & 4096 \\
        Learning rate schedule & Cosine decay\\
        Warmup epochs & 20 \\
        Training epochs & 200 \\
        Augmentation & RandAug (9, 0.5)\\ 
        Label smoothing & 0.1 \\
        Mixup & 0.8 \\
        Cutmix & 1.0 \\
    \bottomrule
    \end{tabular}
    \captionof{table}{\textbf{Training settings for ImageNet.}}
    \label{tab:training—settings_imagenet}
    \end{minipage}
\end{figure*}

\section{Computational cost}
\label{app:computational_cost}
Our experiments were conducted on RTX 3090 and V100 GPUs. The detailed computational costs are presented in Tab.~\ref{tab:cost_time}. Training epochs were set to 50 for CIFAR-10 and 200 for ImageNet. The number of epochs trained on CIFAR-10 is relatively low because we use a pre-trained model to initialize our classifier, enabling faster convergence.

\begin{table*}[t!]
\vskip 0.15in
\small
\centering
\begin{tabular}{cccccc}
\toprule
\textbf{Model} & \textbf{Combinations} & \textbf{Epochs} & \textbf{GPU-type} & \textbf{GPU-nums} & \textbf{Hours} \\
\midrule
ResNet-50 & \textit{C}, \textit{U} & 50 & 3090 & 4 & 2 \\
ViT-S & \textit{C}, \textit{U} & 50 & 3090 & 4 & 2 \\
ConvNeXt-T & \textit{C}, \textit{U} & 50 & 3090 & 4 & 2 \\
\midrule
ResNet-50 & \textit{I}-256, \textit{U}-H/2  & 200 & V100 & 8 & 6 \\
ViT-S & \textit{I}-256, \textit{U}-H/2 & 200 & V100 & 8 & 9 \\
ConvNeXt-T & \textit{I}-256, \textit{U}-H/2 & 200 & V100 & 8 & 8 \\
\midrule
ResNet-50 & \textit{I}-512, \textit{E}2-XXL & 200 & V100 & 8 & 9 \\
ViT-S & \textit{I}-512, \textit{E}2-XXL & 200 & V100 & 8 & 11 \\
ConvNeXt-T & \textit{I}-512, \textit{E}2-XXL & 200 & V100 & 8 & 10\\
\bottomrule
\end{tabular}
\caption{\label{tab:cost_time} \textbf{Training time of classifiers}.}
\end{table*}

\begin{table*}[t]
    \small
    \centering
    \begin{tabular}{ccccc}
    \toprule
        \textbf{Real dataset} &
        \textbf{Generative model} & \textbf{FID} &  \textbf{Classifier} & \textbf{Accuracy (\%)} \\
    \midrule
        \multirow{3}{*}{{\emph{I}-256}} & \multirow{3}{*}{{StyleGAN-XL}~\citep{sauer2022stylegan}} & \multirow{3}{*}{2.30} &  ResNet-50 & 99.69 \\
        &  &  & ViT-S & 99.95 \\
        &  &  & ConvNeXt-T & 96.85\\
    \midrule
        \multirow{1}{*}{{\emph{I}-256}} & \multirow{1}{*}{{SiT-XL}~\citep{ma2024sit}} & \multirow{1}{*}{2.06} &  ResNet-50 & 99.87 \\
    \bottomrule
    \end{tabular}
    \caption{\textbf{Binary distribution classification on label-to-image}. All classifiers yield high accuracy on various datasets against strong generative models. FIDs are taken from the corresponding references.} 
    \label{tab:gan_label_to_image}
\end{table*}

\begin{table*}[!ht]
    \vskip 0.10in
    \small
    \centering
    \begin{tabular}{ccccc}
    \toprule
        \textbf{Real dataset} &
        \textbf{Generative model} & \textbf{Model} & \textbf{Pretrained}  & \textbf{Scratch} \\
    \midrule   
        \multirow{3}{*}{\emph{C}} & \multirow{3}{*}{\emph{E}~\citep{karras2022elucidating}} &  ResNet-50 & 96.25 & 56.13 \\
         & & ViT-S & 89.38 & 55.32 \\
         & & ConvNeXt-T & 98.43 & 53.27 \\
    \midrule
        \multirow{3}{*}{\emph{C}} & \multirow{3}{*}{\emph{U}~\citep{bao2023all}} &  ResNet-50 & 99.92 & 56.70 \\
         & & ViT-S & 98.04 & 52.66 \\
         & & ConvNeXt-T & 99.96 & 56.37 \\
    \bottomrule
    \end{tabular}
     \caption{\textbf{Distribution classification accuracy on CIFAR-10.} "Pretrained" indicates that the classifier was initialized with a model pretrained on ImageNet, while "Scratch" indicates that the classifier was trained from scratch.} 
    \label{tab:cifar_10_results}
\end{table*}

\begin{table*}[t]
    \small
    \centering
    \begin{tabular}{ccc}
    \toprule
        \textbf{Training samples} & \textbf{Self-supervised method} & \textbf{Accuracy (\%)} \\
    \midrule
         5k & MAE & 87.77 \\
         5k & MoCo v3 & 90.48 \\
         10k & MAE & 91.67 \\
         10k & MoCo v3 & 90.90 \\
    \bottomrule
    \end{tabular}
    \caption{\textbf{Four-way distribution classification on text-to-image}. Linear classifier on self-supervised features also achieves high accuracy in distinguishing four distributions: COCO~\citep{coco}, Pixart-$\alpha$~\citep{chen2023pixart}, SDXL~\citep{podell2023sdxl}, and Playground-v2.5~\citep{li2024playground}, using only 5k or 10k training samples per dataset. We use ViT-B as the backbone.} 
    \label{tab:self_supervised_text_to_image}
\end{table*}

\section{Additional results}
\subsection{Additional results from other generative models}
\label{app:gan_label_to_image}
As shown in Tab.~\ref{tab:gan_label_to_image}, we conducted experiments using StyleGAN-XL~\citep{sauer2022stylegan}, a state-of-the-art GAN model trained on ImageNet-256, as well as SiT~\citep{ma2024sit}, a flow matching model that extends the applicability of the method. Our results demonstrate that, despite the use of a discriminator during GAN training, the classifier can still easily distinguish between real and generated images. We argue that this is because the discriminator is trained jointly with the generator. During training, the generated data seen by the discriminator comes from a continuously evolving distribution, as the generator improves with each iteration. However, when using a classifier to distinguish between real and generated distributions, the generated distribution remains fixed. 

Our experiments confirm that images generated by GANs can also be readily distinguished from real images using a classifier, despite GANs' adversarial training approach. We have not yet conducted experiments to explore this phenomenon in other generative frameworks such as Masked Image Generation models~\citep{chang2022maskgit,you2025effective} or Autoregressive models~\citep{sun2024autoregressive, tian2024visual}, which remains an interesting direction for future work.

\subsection{Results of CIFAR-10}
\label{app:result_of_cifar}
In order to ensure the completeness of the experiment, we are here to present the result of CIFAR-10 trained from scratch. We present the results in Tab.~\ref{tab:cifar_10_results}. If there is no prior knowledge, classifiers struggle to distinguish between real and generated data in datasets with low resolution, such as CIFAR-10 (i.e., 32x32). However, the use of a pre-trained model allows the features learned from the ImageNet dataset to aid in differentiating between real and generated images in CIFAR-10.

\subsection{Self-supervised classifiers for Text-to-Image distribution classification}
\label{app:self_supervised_text_to_image}
As shown in Tab.~\ref{tab:self_supervised_text_to_image}, we report the distribution classification accuracy of self-supervised classifiers in text-to-image scenarios. Notably, these classifiers achieve high accuracy in distinguishing between different text-to-image models. This suggests that there are significant differences between text-to-image generative models, allowing even self-supervised classifiers to easily distinguish them.

\subsection{Visualization of crops}
\begin{figure*}[t]
    \centering
    \includegraphics[width=0.5\linewidth]{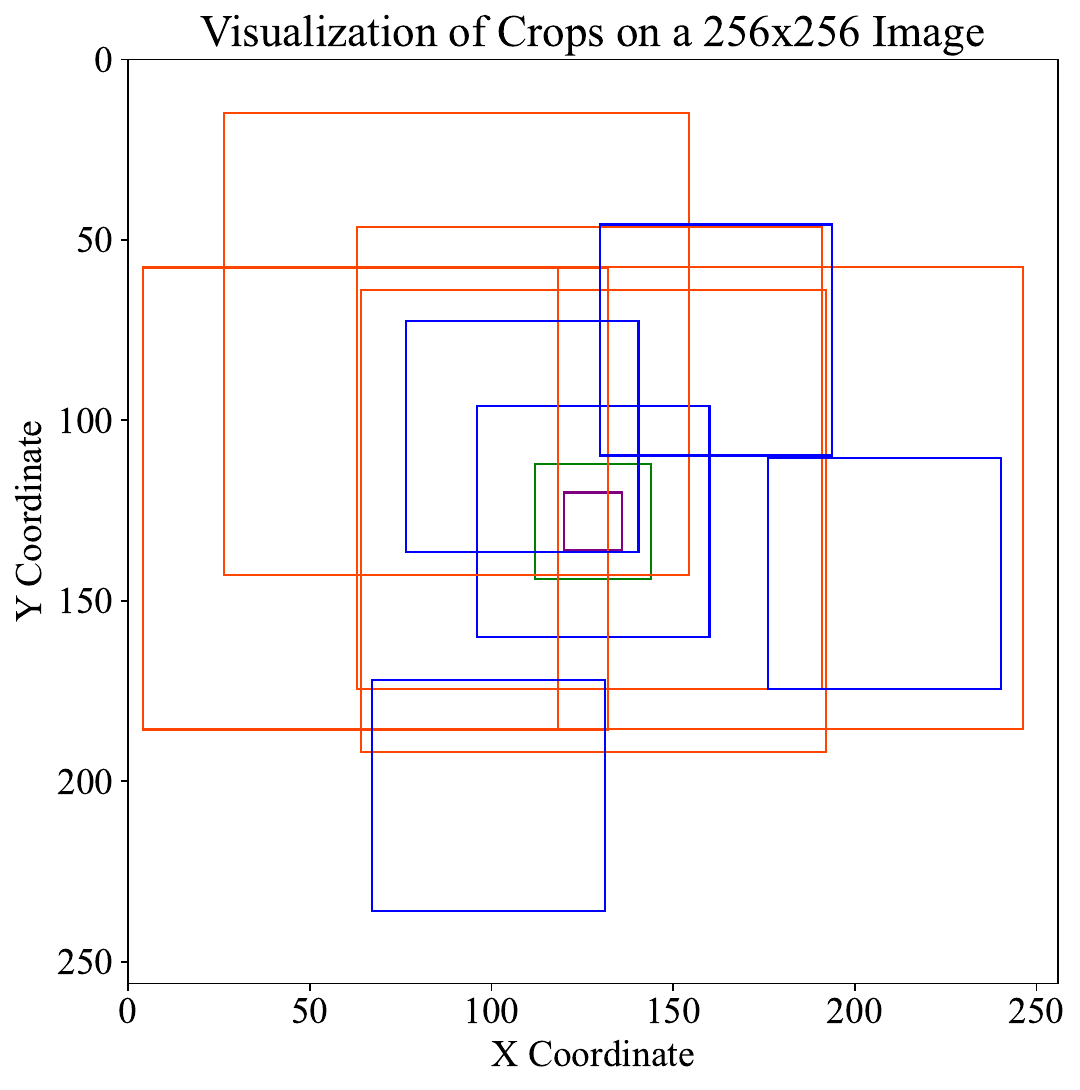}
    \caption{\textbf{Visualization of crops.} With each resolution represented by a different color.}
    \label{fig:crop_visualization}
\end{figure*}

As shown in Fig.~\ref{fig:crop_visualization}, we present the visualization of crops mentioned in Sec.~\ref{sec:evaluate_alignment}.

\subsection{Frequency analysis on the combination of EDM2-XS and EDM2-XXL}
\label{app:frequency_analysis_edm2xs_xxl}
\begin{figure*}
    \centering
    \includegraphics[width=1\linewidth]{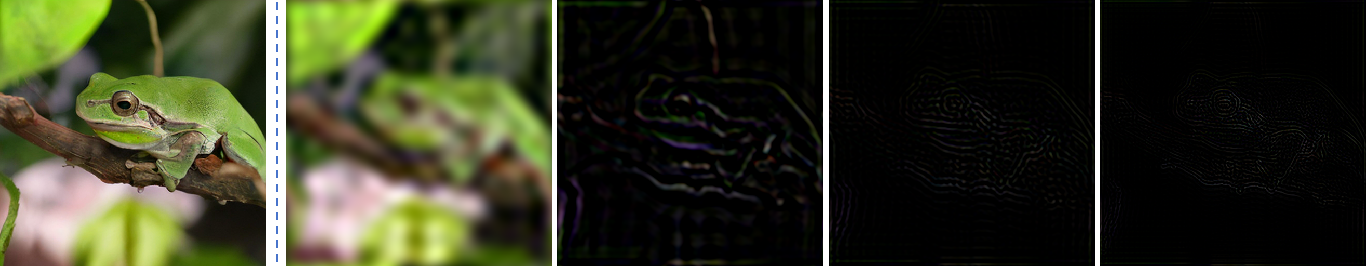}
    \caption{\textbf{Visualization of frequency domain processing of EDM2-XXL.} The image shows the results after applying a band-pass filter to EDM2-XXL with band thresholds of 0-10, 10-30, 30-50, and 50-100, from left to right.}
    \label{fig:frequency_process_generated}
\end{figure*}

\begin{figure}
    \centering
    \includegraphics[width=\linewidth]{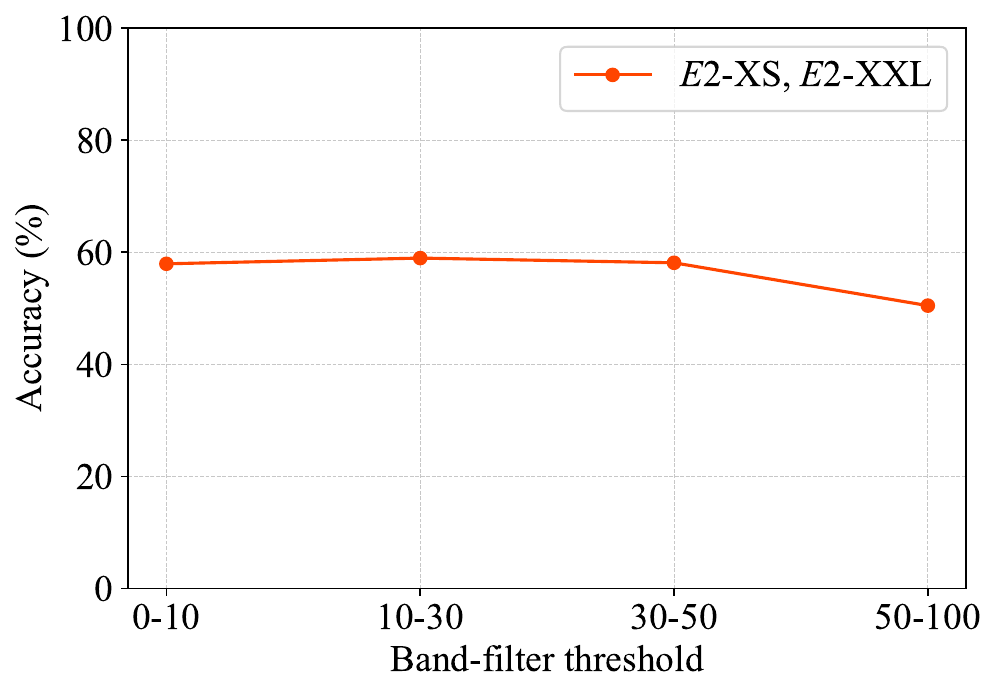}
    \caption{\textbf{Accuracy vs. band-frequency filter threshold}}
    \label{fig:model_accuracy_vs_band_filter_threshold_xs_xxl}
\end{figure}

We preprocess the generated images using band-pass filters as defined in Sec.~\ref{sec:frequency_filter}, with four threshold intervals: 0-10, 10-30, 30-50, and 50-100. An example of original and processed EDM2-XXL generated images is shown in Fig.~\ref{fig:frequency_process_generated}. As shown in Tab.~\ref{tab:frequency_analysis_generated} and Fig.~\ref{fig:model_accuracy_vs_band_filter_threshold_xs_xxl}, for EDM2-XS and EDM2-XXL, the smallest and largest models in the EDM2 family, classifiers' accuracy approaches random guessing (around 50\%) across different threshold intervals. This indicates that for models within the same diffusion model family, which share inductive biases but differ in visual quality, classifiers unable to distinguish between them based on any specific frequency band.

\begin{table}[!ht]
    \vskip 0.10in
    \small
    \centering
    \begin{tabular}{cccc}
    \toprule
        \textbf{Combinations} & \textbf{Classifier} & \textbf{Intervals} & \textbf{Accuracy} \\ 
    \midrule
        \multirow{4}{*}{\emph{E}2-XS, \emph{E}2-XXL} & \multirow{4}{*}{ResNet-50} & 0-10 & 57.94 \\ 
        & & 10-30 & 58.96 \\
        & & 30-50 & 58.12 \\
        & & 50-100 & 50.48 \\ 
    \bottomrule
    \end{tabular}
     \caption{\textbf{Classification accuracy} of ResNet-50 on the combination of EDM2-XS and EDM2-XXL after applying band-pass filters.} 
    \label{tab:frequency_analysis_generated}
\end{table}

\subsection{User study} 
\label{app:user_study}
\begin{figure*}[t]
    \centering
    \begin{subfigure}[b]{0.47\linewidth}
        \includegraphics[width=\linewidth]{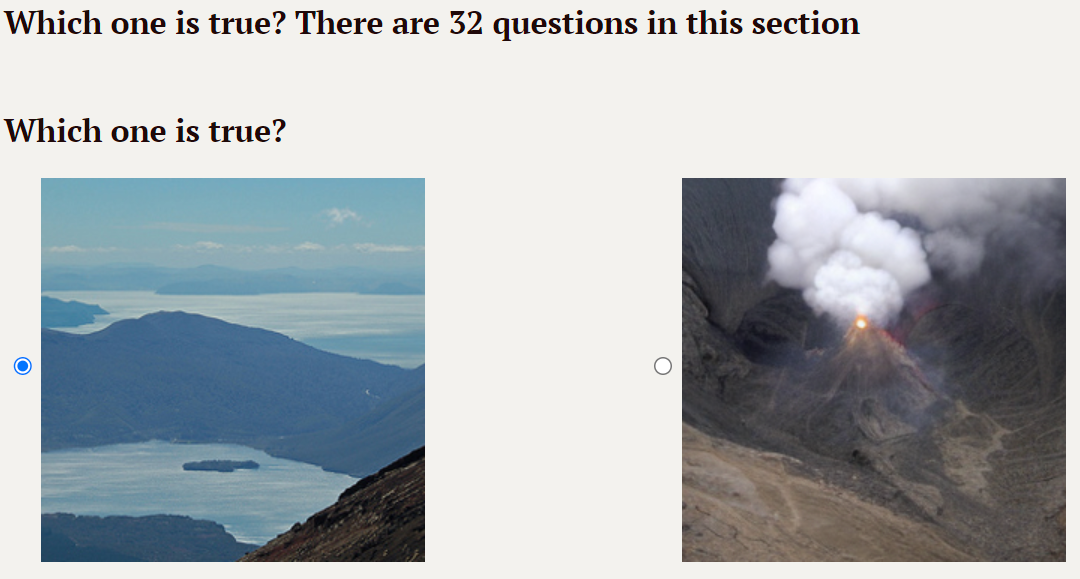}
        \caption{\textit{I}-256 vs. \textit{U}-ViT-H/2}
        \label{fig:user_1}
    \end{subfigure}
    \hfill
    \begin{subfigure}[b]{0.47\linewidth}
        \includegraphics[width=\linewidth]{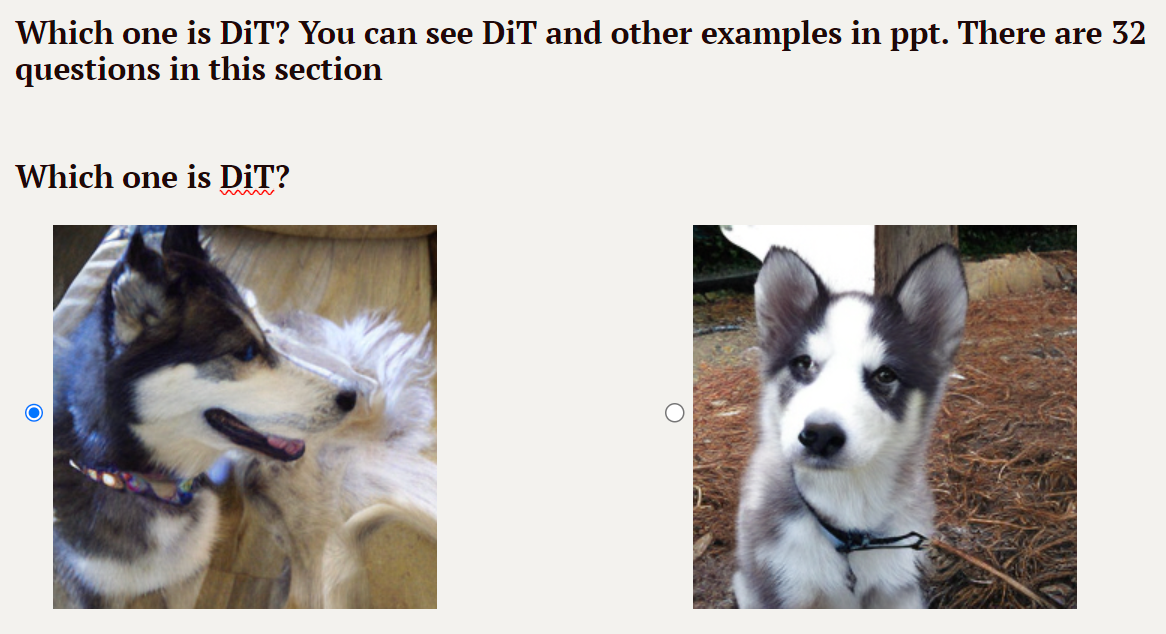}
        \caption{\textit{U}-ViT-H/2 vs. \textit{D}}    
        \label{fig:user_2}
    \end{subfigure}
    \hfill
    \begin{subfigure}[b]{0.6\linewidth}
        \includegraphics[width=\linewidth]{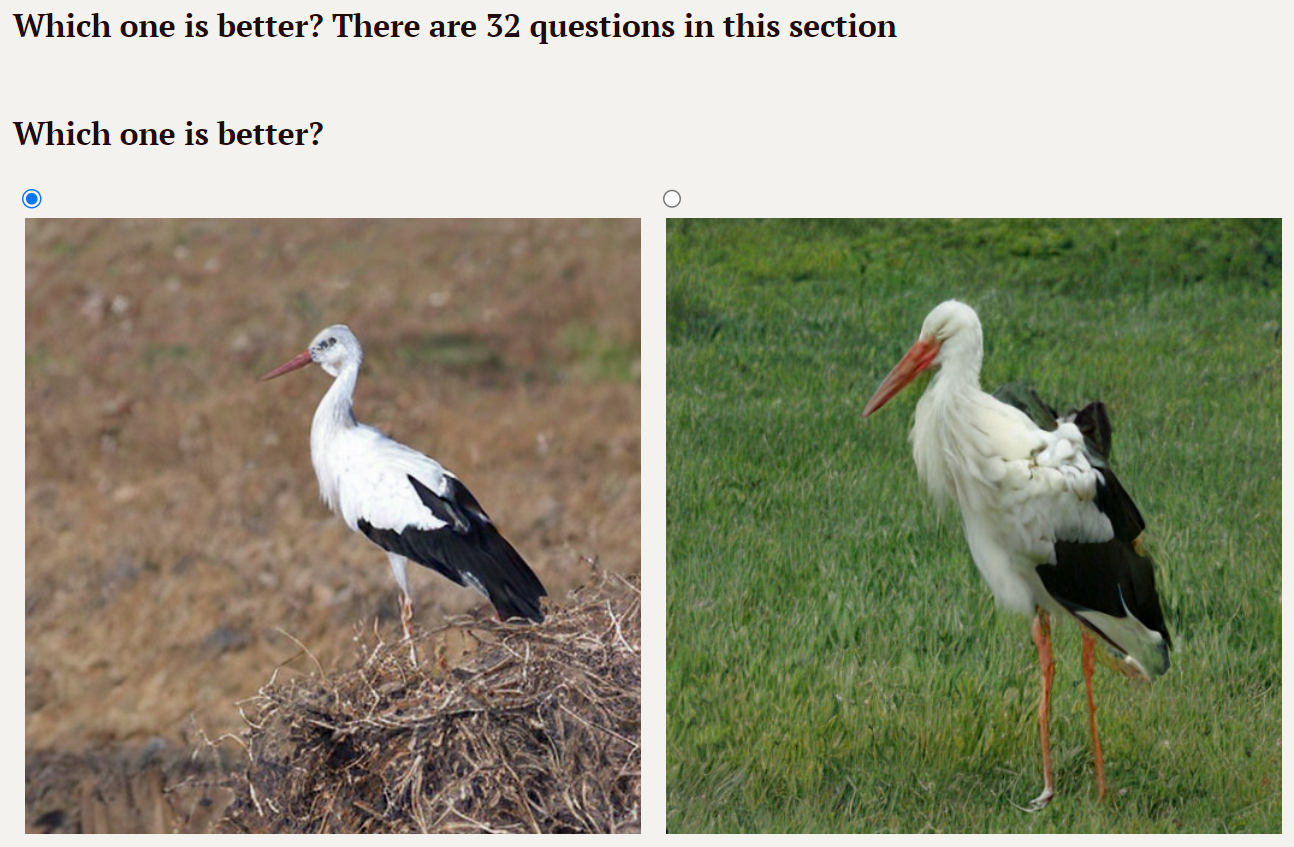}
        \caption{\textit{E}2-XS vs. \textit{E}2-XXL}        
        \label{fig:user_3}
    \end{subfigure}
    \caption{\textbf{Screenshot of user study.} Participants are asked to distinguish generated distributions from real ones and to classify which diffusion model generated a given image. Each group of experiments is illustrated with a separate example here. In each set of experiments, we also randomized the order to prevent examples from the same set from influencing each other.}
    \label{fig:user_study_appendix}
\end{figure*}

Fig.~\ref{fig:user_study_appendix} shows a screenshot of the interaction interface used in our user study. The study involved nineteen participants, and we designed three groups of experiments, each requiring participants to classify 32 pairs of images. In the first set, participants distinguished between generated and real images, with real images sourced from ImageNet-256 and generated images from U-ViT-H/2. In the second set, they classified images between two diffusion models with similar performance: DiT-XL/2 and U-ViT-H/2. For the final set, participants evaluated images from EDM2-XS and EDM2-XXL, which share the same training methodology but differ in parameter count, resulting in different FID scores and visual quality. In the first set, participants were asked to identify the real images. In the second, they were tasked with identifying DiT images, and reference images from DiT and U-ViT were provided during the test. In the third experiment, participants judged which images were of higher quality. All nineteen participants were graduate students with substantial experience in machine learning. They were allowed to zoom in on the images during the experiment, which was conducted on 27-inch 4K displays. All participants had corrected vision of 1.0 (standard normal vision), and their ages ranged from 22 to 26. Each participant completed the experiment within an hour and was compensated \$10.

\section{Binary classification as a measure of distribution distance}
\label{app:binary_classification_distance}
We employ a classifier \( C(x) \) to distinguish between the real data distribution \( p_{\text{data}}(x) \) and the generated data distribution \( p_g(x) \). By training \( C(x) \) using the binary cross-entropy loss:
\begin{align}
L(C) = -E_{x \sim p_{\text{data}}(x)}[\log(C(x))] - E_{x \sim p_g(x)}[\log(1 - C(x))],
\end{align}
The optimal classifier that minimizes this loss is:
\begin{align}
C^*(x) = \frac{p_{\text{data}}(x)}{p_{\text{data}}(x) + p_g(x)}.
\end{align}
Substituting \( C^*(x) \) back into the loss function yields:
\begin{align}
L(C^*) = -\log(4) + 2\, \text{JSD}(p_{\text{data}}(x) \| p_g(x)),
\end{align}
where \( \text{JSD} \) denotes the Jensen-Shannon Divergence—a direct measure of the distance between the two distributions.

In our paper, we use classification accuracy to evaluate how well the classifier distinguishes between real and generated data because it provides an intuitive and interpretable metric. Although accuracy is non-differentiable and unsuitable for direct optimization, training the classifier with the binary cross-entropy loss—a convex surrogate—often leads to improved accuracy. This correlation suggests that accuracy can serve as a proxy for changes in the loss function, reflecting the distance between the generated and real data distributions.

\begin{figure*}[t]
    \centering
    \begin{subfigure}[b]{0.475\linewidth}
        \includegraphics[width=\linewidth]{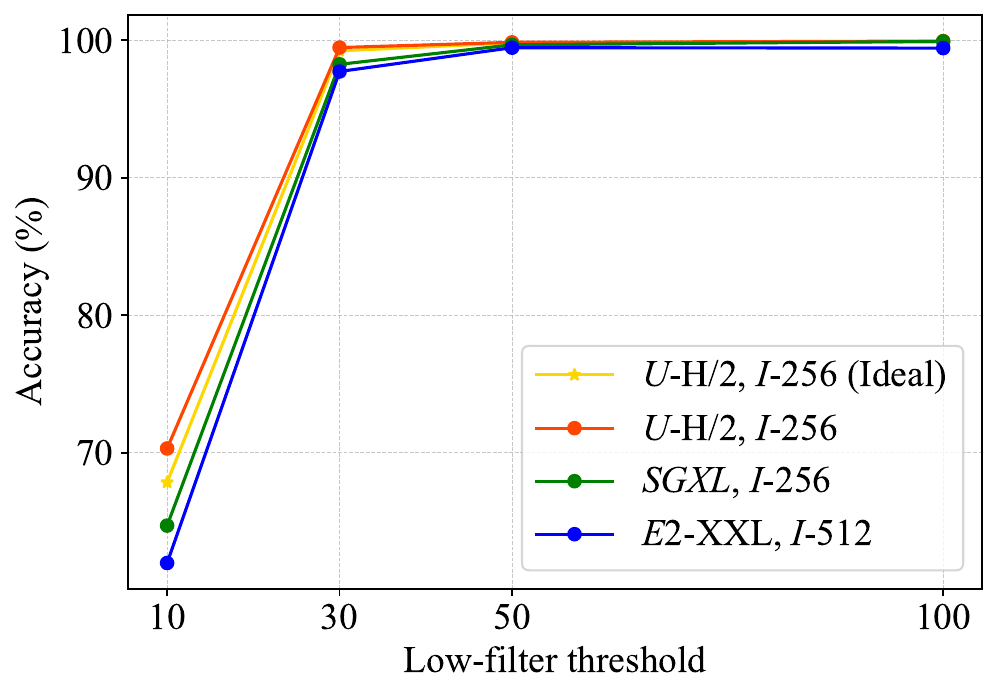}
        \caption{Accuracy vs. low-frequency filter threshold}
        \label{fig:resnet50_low_filter_choice_experiment}
    \end{subfigure}
    \hfill
    \begin{subfigure}[b]{0.475\linewidth}
        \includegraphics[width=\linewidth]{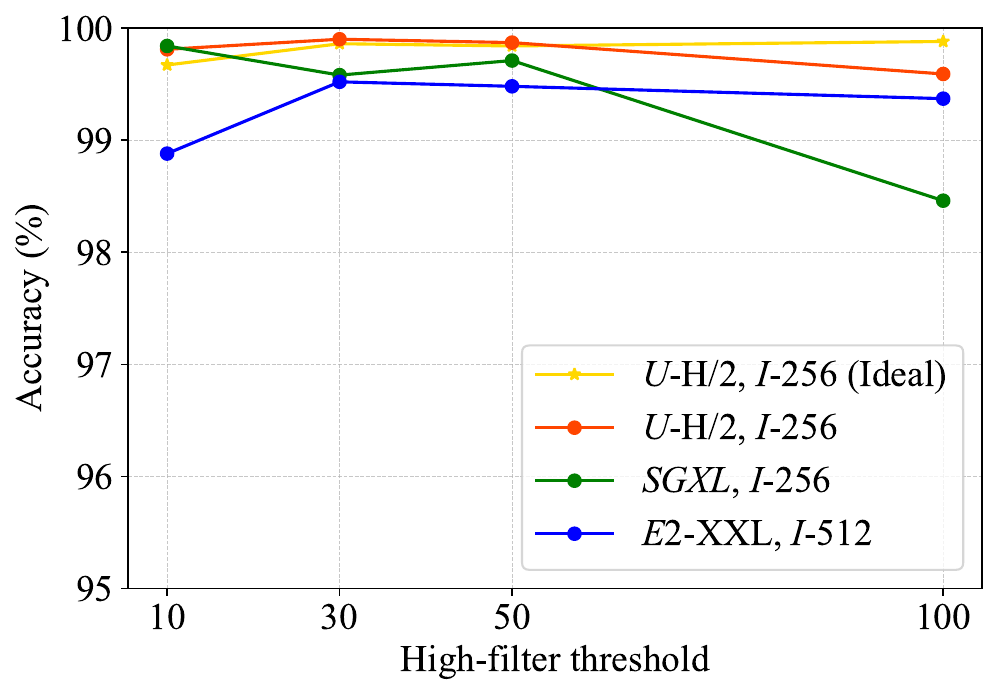}
        \caption{Accuracy vs. high-frequency filter threshold}        \label{fig:resnet50_high_filter_choice_experiment}
    \end{subfigure}
    \caption{Comparison of model accuracy across different filter thresholds using rectangular and circular filters.}
    \label{fig:resnet50_filter_choice_experiment}
\end{figure*}

\section{Relationship between distribution classification and FID}
\label{app:relationship_fid}
The Fréchet Inception Distance (FID) is a widely used metric for evaluating the quality of generative models. It relies on feature extraction networks trained on datasets such as ImageNet and assumes that the extracted feature vectors follow a multivariate Gaussian distribution. FID calculates the Fréchet distance between these Gaussian distributions to measure discrepancies between the real and generated data. Meanwhile, as noted by \citet{kynkaanniemi2022role}, FID can decrease simply by aligning the histograms of top-N classifications, without necessarily improving the perceptual quality of the generated images. Additionally, recent works like \citet{karras2023analyzing} and \citet{tian2024visual} report FID scores close to those of the ImageNet validation set, suggesting limitations in FID's sensitivity to certain distribution differences.

In contrast, our classifier-based approach offers a more direct measure of the distance between distributions without relying on the Gaussian assumptions inherent in FID. By training a classifier to distinguish between real and generated data, we obtain an intuitive and interpretable metric that reflects the actual distribution differences. This method complements commonly used metrics such as FID and Inception Score (IS), providing an alternative perspective on evaluating generative models.

\section{Comparison of rectangular and circular filters}
\label{app:choice_filter}
Rectangular and circular filters are common techniques in image filtering. In this paper, we chose to implement a rectangular filter following the official FreeU implementation~\citep{si2023freeu}, due to its simplicity and computational efficiency. For comparison, Fig.~\ref{fig:resnet50_filter_choice_experiment} presents initial results using a circular mask, denoted as U-H/2 and I-256 (Ideal). The results from both implementations are similar, and we chose to use the rectangular filter for its computational efficiency.

\end{document}